\documentclass{article} % For LaTeX2e
\usepackage{collas2023_conference,times}
\usepackage{easyReview}

\usepackage{amsmath,amsfonts,bm}

\def\eqref#1{equation~\ref{#1}}
\def\1{\bm{1}}

\DeclareMathAlphabet{\mathsfit}{\encodingdefault}{\sfdefault}{m}{sl}
\SetMathAlphabet{\mathsfit}{bold}{\encodingdefault}{\sfdefault}{bx}{n}

\usepackage{hyperref}

\usepackage{graphicx}
\usepackage{amsmath}
\usepackage{amstext}
\usepackage{amsfonts}
\usepackage{amssymb}
\usepackage{booktabs}
\usepackage{comment}
\usepackage{algorithm}
\usepackage{subcaption}
\usepackage{algpseudocode}
\usepackage{todonotes}
\usepackage{wrapfig}
\usepackage{multirow}
\usepackage{colortbl}
\usepackage{makecell}
\usepackage{amssymb}

\usepackage{colortbl}
\usepackage{xcolor}
\usepackage{soul}

\hypersetup{
    colorlinks=true,
    linkcolor=red,
    filecolor=magenta,
    urlcolor=blue,
    citecolor=purple,
    pdftitle={Overleaf Example},
    pdfpagemode=FullScreen,
    }

\title{Partial Hypernetworks for Continual Learning}

\author{
    \hspace{2.5cm} Hamed Hemati\textsuperscript{1}, Vincenzo Lomonaco\textsuperscript{2}, Davide Bacciu\textsuperscript{2}, Damian Borth\textsuperscript{1}
    \\
    \\
    \hspace{0.6cm} hamed.hemati@unisg.ch, vincenzo.lomonaco@unipi.it, davide.bacciu@unipi.it, damian.borth@unisg.ch
    \\
    \\
    \hspace{3.2cm}\textsuperscript{1}University of St. Gallen, Switzerland  \hspace{0.2cm} \textsuperscript{2} University of Pisa, Italy 
}

\collasfinalcopy % Uncomment for camera-ready version, but NOT for submission.

\begin{document}

\maketitle

\begin{abstract}
Hypernetworks mitigate forgetting in continual learning (CL) by generating task-dependent weights and penalizing weight changes at a meta-model level. Unfortunately, generating all weights is not only computationally expensive for larger architectures, but also, it is not well understood whether generating all model weights is necessary. Inspired by latent replay methods in CL, we propose partial weight generation for the final layers of a model using hypernetworks while freezing the initial layers. With this objective, we first answer the question of how many layers can be frozen without compromising the final performance. Through several experiments, we empirically show that the number of layers that can be frozen is proportional to the distributional similarity in the CL stream. Then, to demonstrate the effectiveness of hypernetworks, we show that noisy streams can significantly impact the performance of latent replay methods, leading to increased forgetting when features from noisy experiences are replayed with old samples. In contrast, partial hypernetworks are more robust to noise by maintaining accuracy on previous experiences. Finally, we conduct experiments on the split CIFAR-100 and TinyImagenet benchmarks and compare different versions of partial hypernetworks to latent replay methods. We conclude that partial weight generation using hypernetworks is a promising solution to the problem of forgetting in neural networks. It can provide an effective balance between computation and final test accuracy in CL streams.

\end{abstract}

\section{Introduction}
One of the core assumptions in statistical learning is that the data samples from a given dataset are independent and identically distributed (IID) \cite{hastie2009elements}. This assumption requires the model to be trained and evaluated on a fixed data distribution. However, there are different cases in which this assumption can be violated due to changes in the data distribution, either in training or in the test phase. Such cases include multi-task learning, adversarial examples, and continual learning \cite{murphy2022probabilistic}.

Continual learning focuses on a model's ability to learn from new experiences over time \cite{thrun1995lifelong} when exposed to a data stream. This capability is particularly useful in real-world applications across different domains, such as image understanding \cite{parisi2019continual, de2021continual}, natural language processing \cite{biesialska2020continual}, reinforcement learning \cite{khetarpal2020towards}, and robotics \cite{lesort2020continual}. However, there are multiple challenges to learning from a stream of experiences. One such challenge is catastrophic forgetting (CF) \cite{french1999catastrophic}, which occurs when information from older experiences is partially or completely replaced by information from newer experiences. This can be problematic when models are deployed for real-world scenarios, as the model must adapt its knowledge to new situations while preserving its previously learned information.

Rehearsal-based methods are a well-known category of strategies that are highly regarded for their simplicity and efficiency in mitigating CF by replaying exemplars from previously encountered experiences. These exemplars may take on a variety of forms, including raw samples from the dataset \cite{rolnick2019experience}, synthesized samples generated via a generative model \cite{shin2017continual, graffieti2022generative}, or in latent replay (LR) form where features are obtained from a frozen feature extractor \cite{ostapenko2022foundational, demosthenous2021continual}. In LR, the number of frozen layers after the pre-training phase can significantly impact the model's ability to learn from subsequent experiences. For instance, freezing all model layers, except for the classification layer, is reasonable only when there are close affinities between upcoming experiences and the one used for pre-training, as the feature extractor directly influences the quality of features.

Given these points, critical questions need to be addressed in LR methods with frozen feature extractors. The first question is \textit{How many layers can we freeze without losing generalizability?} And the second question is \textit{How sensitive are these approaches to distributional changes?} The first question can be partly addressed by allocating larger capacity to the model layers succeeding the frozen feature extractor by finding a sound trade-off between the number of frozen and non-frozen layers. The latter, however, requires a more careful examination of how replaying features extracted from a fixed feature extractor affects the model's learning across different distributions in terms of negative interference, an issue also studied by \cite{wang2021afec}.

One possible approach to address multi-task learning problems with high negative interference between tasks is to use Hypernetworks (HNs) \cite{ha2016hypernetworks}. HNs are meta-models that produce the weights of another network, commonly known as the "main model". The weights can be generated in response to different types of signals, such as the task ID in a multi-task learning dataset or the input itself to create input-specific weights. HNs offer several advantages, including model compression \cite{nguyen2021fast}, neural architecture search \cite{brock2017smash}, and continual learning \cite{von2019continual}. By capturing the shared structure between tasks, HNs can improve generalization to new tasks \cite{beck2022hypernetworks} through adaptive weight generation. In continual learning, HNs have been demonstrated to mitigate forgetting significantly \cite{von2019continual}, although training can be computationally expensive \cite{cubelier2022master}. However, HNs can be used in a more efficient way. For example, \cite{ustun2022hyper} proposed generating only the weights of adaptive layers in a neural network. This approach can potentially reduce the computational costs associated with HNs while still leveraging their benefits.

We propose to use partial hypernetworks to generate the weights of non-frozen layers in a model as an alternative to replaying stored features. This idea is inspired by recent work on LR with frozen layers and motivated by the observation that lower layer representations remain stable after the first experience \cite{ramasesh2020anatomy}. Using HNs for the final layers provides advantages over regular LR, and standard HN approaches. First, we avoid storing exemplars and instead perform replay at the meta-model level by keeping the converged version of the hypernetwork from the previous step, which is typically smaller than the main model. Second, we avoid generating the weights of layers shared among experiences, which can speed up model training while achieving comparable performance to regular HNs. We comprehensively analyze different hypernetwork architectures and evaluate their performance on various benchmark datasets to compare these models with their LR and standard HN counterparts. Below, we summarize our contributions:

\begin{itemize}
    \item We propose a more efficient solution for hypernetworks by combining the concept of a frozen feature extractor with hypernetworks.
    \item We demonstrate that generating partial weights using hypernetworks can achieve comparable performance under specific conditions.
    \item We offer insights into when these models can be a superior alternative to conventional latent replay techniques.
\end{itemize}

\begin{figure}[t]
    \centering
    \includegraphics[width=0.99\textwidth]{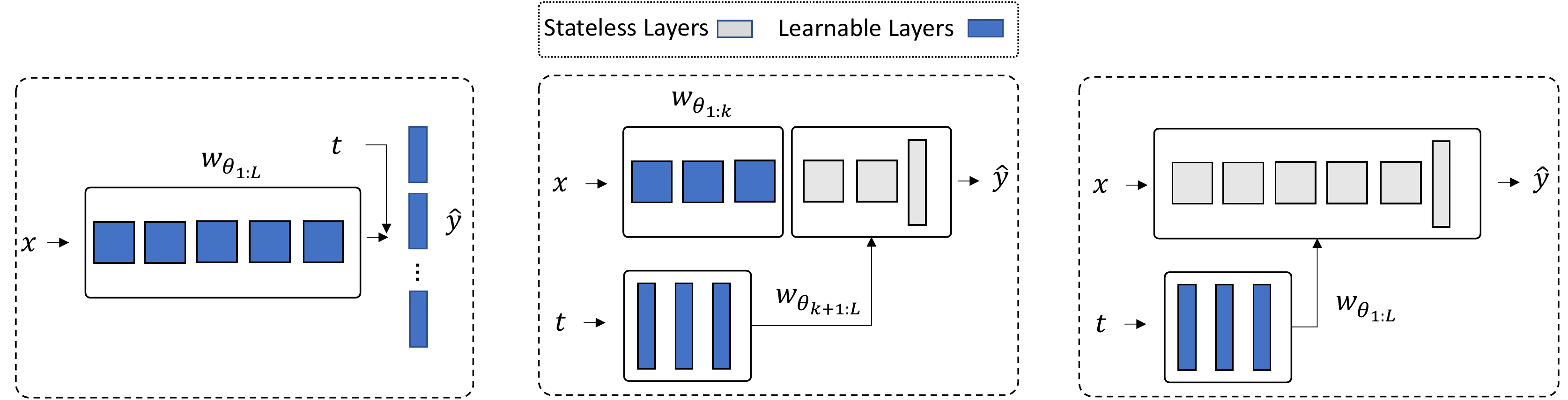}
    \caption{Left: a regular network with multiple heads. Middle: a partial hypernetwork that generates the weights of the final layers of the main model. Right: a hypernetwork that generates all weights of the main model.}
    \label{fig:network_types}
\end{figure}

\section{Problem Definition}

In a standard supervised classification problem using neural networks, we assume we have a neural network $\mathbf{f}_{ \theta}(x)$ with learnable parameters $\theta$. Given a dataset $\mathbf{D}=\{(x,y); x \in \mathbf{X}, y \in \mathbf{Y}\}$  with underlying distribution $\mathcal{D}$ where $\mathbf{X} \subseteq \mathbb{R}^{d}$ and $\mathbf{Y}=\{1, 2, \ldots C \}$ represent the input features with dimensionality $d$ and output labels in the category space, respectively; the objective is to learn a mapping $\mathbf{f}_{\theta}: \mathbf{X} \rightarrow \mathbf{Y}$ that has a low empirical risk over the dataset samples. In other words, we seek to minimize the difference between the mappings and ground-truth samples by finding $\theta^* = \arg\min_{\theta} L$, where $L=\sum_{i=0}^{N}{CE(\mathbf{f}_{\theta}(x), y)}$ denotes the empirical risk, and $CE$ is the function that calculates the cross-entropy loss, with samples drawn from  $\mathcal{D}$ in an IID fashion. We obtain a solution with a predetermined error threshold using the stochastic gradient descent (SGD) algorithm.

\paragraph{Continual Learning} in contrast to the standard supervised learning, is a setting in which the entire dataset $\mathbf{D}$ is not available at once. Instead, data becomes available over time in the form of a data stream. Specifically, the model is exposed to a sequence of experiences $\mathcal{S} = \{\mathbf{e}_1, \mathbf{e}_2, ..., \mathbf{e}_N\}$, where each $\mathbf{e}_i, 0 < i < N$ represents an individual experience that contains a train and test set, i.e., $\mathbf{e}_i=\{\mathbf{D}^{(i)}_{tr}, \mathbf{D}^{(i)}_{te}, t^{(i)} \}$ with $\mathbf{D}^*_{tr}$, $\mathbf{D}^*_{te}$ and $t^{(i)}$ being the train set, test set and the task id of the experience. The datasets in each experience can be drawn from the same distribution or have different distributions.

A \textit{strategy} refers to an algorithm that updates the model $\mathbf{f_{\theta}}$ over a sequence of experiences. When a strategy goes through each experience, it does so one after the other and does not revisit any previously observed experience. Once the strategy observes the $N^{th}$ experience $\mathbf{e}_N$, it obtains the corresponding converged parameters $\theta^{*}_{N}$. After each experience, the model is evaluated by calculating the Average Classification Accuracy (ACA) over the test set of all the experiences observed so far, i.e., $\text{ACA}=\sum_{i=1}^{N}\text{Acc}(\mathbf{f}_{\theta_{T}^{*}}, \mathbf{D}_{te}^{(i)})$ with $\text{Acc}(\mathbf{f}, \mathbf{D})$ being the accuracy of a function $\mathbf{f}$ over dataset $\mathbf{D}$. Strategies are typically compared in terms of their ability to preserve knowledge, which refers to the degree to which the model's performance is affected after going through all experiences in the stream.

The type of \textit{scenario} in CL determines what kind of information a new experience provides and how the input and output domains change over time. The most commonly encountered scenarios are domain-incremental, task-incremental, and class-incremental \cite{van2019three}. In the domain-incremental scenario, the category space remains constant for all experiences, while only the "domain" of the mapping changes. In both task-incremental and class-incremental scenarios, the category space changes as new classes are introduced. However, in the task-incremental scenario, the experience indicator is given during test time, while in the class-incremental scenario, the model is not presented with any form of task indicator at test time. It is possible to formulate the class-incremental scenario as a task-incremental learning problem with an additional task inference step as theoretically formulated in \cite{kim2022theoretical}. In this case, the mapping $\mathbf{f}$ becomes $\mathbf{f}: \mathbf{X} \times \mathbf{T} \rightarrow \mathbf{Y}$ where in the case of class-incremental, $\mathbf{T}$ has to be inferred. In our experimental setting, similar to \cite{von2019continual}, we assume that the task ID is either given or inferred, which converts our problem into a task-incremental learning problem.

\begin{figure}[h]
    \centering
    \includegraphics[width=0.7\textwidth]{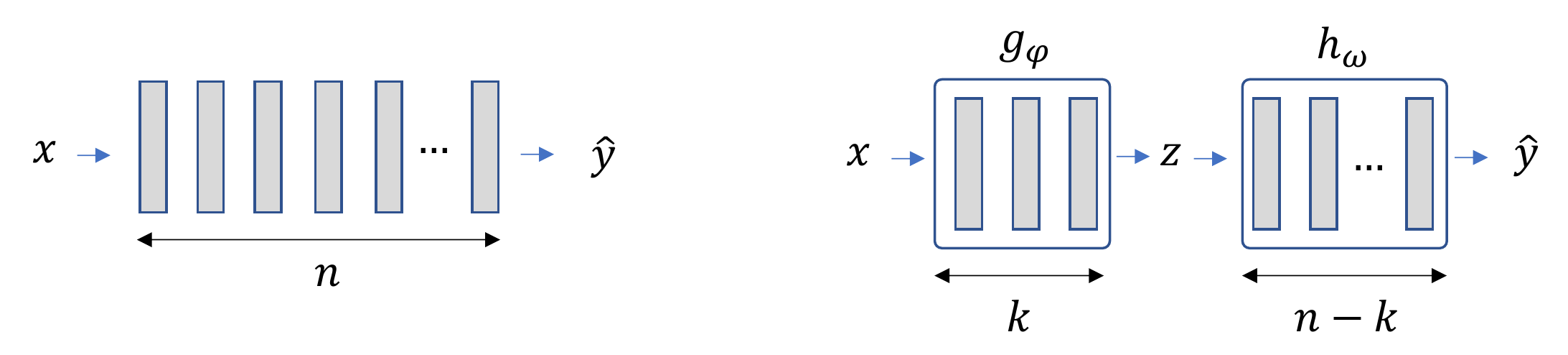}
    \caption{Decomposition of a neural network with $n$ layers into two parts with decomposition depth $k$.}
    \label{fig:decomp}.
\end{figure}

\paragraph{Model Decomposition} is an operation that converts a function $\mathbf{f}$ into a composite function $\mathbf{f}_{P}(\mathbf{f}_{P-1}(...(\mathbf{f}_{1}(x))))$, where $P$ is the total number of consecutive chunks that result from splitting the neural network into $P$ parts. By decomposing a function into simpler parts, we can gain insights into its behavior and use it to solve problems more flexibly. For example, we often decompose functions in calculus to apply techniques like the chain rule or integration by substitution. A common form of decomposition in neural networks is to break them down into two simpler neural networks. In the case of LR with frozen layers, we decompose neural networks into two parts, where the parameters in the first half's layers are frozen and used for feature extraction. In our problem setting, we perform depth-wise decomposition by always splitting a neural network based on a given depth. For depth-specific decomposition, we consider a model $\mathbf{f}_{\theta, l}$ with $l \geq 2$, where $l$ indicates the number of layers, and we define a decomposition for depth $k$ as below:
\begin{equation}
    O_{dec}(\mathbf{f}_{\theta}, k) = \mathbf{f}_{\theta^{1:k}}(\mathbf{f}_{\theta^{k:n}}(x))
    \label{eq_decomp_1}
\end{equation}

$O_{dec}$ denotes the decomposition operation , which is applied for depth $k$, and $\theta^{a:b}$ denotes the subset of parameters from consecutive layers $a$ to $b$. In the case of LR, $k$ is a hyper-parameter that should be set based on the model architecture, computational limitations, and the level of shared structure across tasks. An alternative notation to represent the model segments obtained after decomposition is $\mathbf{f}_{\theta^{1:k}} = \mathbf{g}_\phi$ and $\mathbf{f}_{\theta^{k:l}} = \mathbf{h}_\omega$. With this notation, the inference is performed by first computing $z= \mathbf{g}_{\phi}(x)$ and then using $\mathbf{h}_{\omega}$ to compute $\hat{y} = h_{\omega}(z)$. Figure \ref{fig:decomp} provides an example of such a decomposition.

\paragraph{Partial Weight Generation} is a technique utilized for task-conditional prediction in multi-task or continual learning problems. The primary goal is to generate task-specific weights for particular layers of a model $\mathbf{f}$, and HNs are mainly employed for this purpose. The hypernetwork $\mathbf{H}_{\Psi}$ generates the parameters $\theta$ of model $\mathbf{f}$ before making prediction. When predicting input $x$, $\mathbf{H}_{\Psi}$ first generates the weights conditioned on an additional conditioning vector $t$, which can be the input $x$ itself or a task-dependent indicator such as the task ID. This weight generation process can also be considered an adaptive approach for prediction since the predictor's weights are initially adapted in a generative way for the given task ID and then used for prediction. In our setting, we always decompose $\mathbf{f}$ into two parts, i.e. $\mathbf{h}_{\omega}(\mathbf{g}\phi(x))$, as shown in Equation \ref{eq_decomp_1} and make prediction as below:

\begin{align}
    w_{\phi} = \mathbf{H}_{\psi}(t) \\
    z = \mathbf{g}_{\phi}(x; w_{\phi}) \\
    \hat{y} = \mathbf{h}_{\omega}(z)
\end{align}

The layers corresponding to the first half are called learnable or \textit{stateful} layers and the layers in the second parts whose weights are generated by the hypernetwork are referred to as \textit{stateless} layers.

\paragraph{Continual Learning with Decomposed Models} exploits a decomposed neural network $\mathbf{f(x)} = \mathbf{h}_{\omega}(\mathbf{g}_\phi(x))$ with decomposition depth $k$ to learn from a stream of experiences with minimal catastrophic forgetting. At each step of training, we are given a batch of data samples $(x, y, t)$ which is then used to update $\mathbf{f}$. For our experimental setup, we assume that we freeze the parameters $\phi$ after the first experience in the stream. Regarding used strategies, we mainly consider two sets of methods: LR and partial HNs. The LR methods are facilitated with an additional buffer $\mathcal{B}$ with a limited capacity which stores features from previous samples extracted from $\mathbf{g}_\phi$. The details of the HN architecture are given in Appendix \ref{appendix_modeldetails}.

\section{Method}
In this section, we describe our method that updates a partial HN over a stream of experiences. We split the method into two phases. In the first phase, the model is trained on the first experience until convergence, then the first part of the model is frozen. From the second experience, the model is updated using a look-ahead mechanism \cite{zhang2019lookahead, gupta2020look, von2019continual}.

At each training step, we are given a batch of data samples $b=(x, y, t)$ where $t$ is the experience ID. We assume $\theta = \{\phi, \omega \}$ constitute the parameters of model $\mathbf{f}$ and the hypernetwork is parameterized by $\psi$. We show the steps of our method in Algorithm \ref{alg:training_alg} in Appendix \ref{appendix_pseudocode}. 

\paragraph{Experience $e_1$: }
In the first experience $e_1$, we train the model with regular SGD updates. Using task indicator $t$, we first infer the parameters $\omega$ of the stateless layers of the main model via the hypernetwork: $\omega=\mathbf{H}_{\psi}(t)$. After generating the stateless weights in the main model, we make a prediction on the batch inputs $x$ to obtain $\hat{y}=\mathbf{h}(z=\mathbf{g}(x;\omega))$. The cross-entropy loss $\mathcal{L}_{CE} = CE(y, \hat{y})$ is used to compute the gradients of learnable (stateful) layers and the hypernetwork:

\begin{align}
    \phi \leftarrow \phi - \alpha * \frac{\partial \mathcal{L}_{CE}}{\partial \phi} \\
    \psi  \leftarrow \psi - \beta * \frac{\partial \mathcal{L}_{CE}}{\partial \psi} 
\end{align}

where $\alpha$ and $\beta$ are the learning rates for the main network and the hypernetwork, accordingly. The same process is repeated until convergence to obtain $H^{*(1)}_{\psi}$.

\paragraph{Experience $e_i, i \geq 2$: } Before starting each new experience $e_i$, we first store a copy of the converged hypernetwork $\mathbf{H}^{*}_{\psi} \leftarrow \mathbf{H}^{*(i-1)}_{\psi}$ from the previous experience. Similar to \cite{von2019continual}, we use $\mathbf{H}^{*}_{\psi}$ to penalize the hypernetwork's outputs for previous experiences:

\begin{align}
    L_{H} = \sum_{j=1}^{i-1} \Vert \mathbf{H}^{*}_{\psi}(i) - \mathbf{H}_{\psi}(i) \Vert_2^{2}
    \label{diffnorm}
\end{align}

A naive way to compute the gradients for each step would be to add the regularization term to the cross-entropy loss directly and backpropagate through the network. This would be problematic on the current batch as it will result in inaccurate gradients. We show a visualization of this effect in Appendix \ref{appendix_lookahead}. To fruther explain this effect, we give an example of why this happens:

Let's assume the model is provided with batch $b=(x, y, t)$. In the first step of the experience, the regularization loss will be zero since $H_{\psi}$ only changes after making an update on the current batch $b$. Therefore the gradients computed via $\mathcal{L}_{H}$ will always be one step behind, and the final gradient obtained for each step will be inaccurate. This may not have a big impact in streams with similar tasks, however in streams with big shifts in the distributions of the tasks, it can results in a noticeable difference. Simply put, we aim to find a way to incorporate the impact of the current update made via $\mathcal{L}_{CE}$ on the hypernetwork's outputs for previous tasks.

One solution is to use a different update method such as look-ahead to also obtain gradients for penalizing the changes of $H_{\psi}$ from the first step of the experience. Look-ahead is an optimization method, where instead of \textit{immediately} updating the model on a given batch of data, it first virtually updates the model for one or more steps, then computes the gradients in the final step of the virtual model, w.r.t. the original parameters. This way, we can compute more accurate gradient $\psi$ by backpropagating through the optimization trajectory to measure the impact of the changes made by the new batch $b$ on the parameters that affect the previous tasks. This process is shown in Figure \ref{fig:lookahead}.

\begin{figure}[t]
  \centering
%   \fbox{\rule{0pt}{2in} \rule{0.9\linewidth}{0pt}}
    \includegraphics[width=0.9\linewidth]{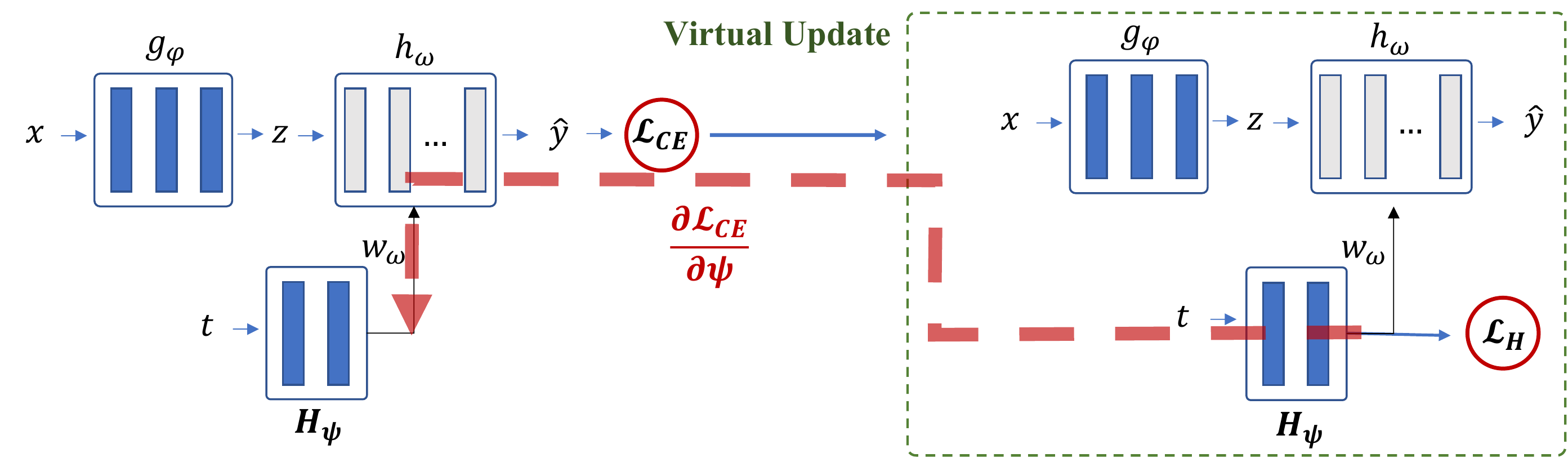}

   \caption{Illustrations of look-ahead gradient computation for the parameters of $\mathbf{H}$. }
   \label{fig:lookahead}
\end{figure}

Thus, the gradient calculation is done in two steps:
% \paragraph{Step 1: } Calculate loss $\mathcal{L}_{CE}$ on the current batch and compute gradients w.r.t. $\phi^{(0)}=\phi$ and $\psi^{(0)}=\psi$:
\paragraph{Step 1: } Calculate loss $\mathcal{L}_{CE}$ on the current batch and compute gradients w.r.t. $\phi^{(0)}=\phi$ and $\psi^{(0)}=\psi$:

\begin{align}
    % \nabla_{\phi} \mathbf{g}_{1} = \frac{\partial \mathcal{L}_{CE}}{\partial \phi^{(0)}} \\
    \nabla_{\psi}\mathbf{H}_{1} = \frac{\partial \mathcal{L}_{CE}}{\partial \psi^{(0)}}
\end{align}

% Use $\nabla_{\phi} \mathbf{g}_{1} $ and $\nabla_{\psi}\mathbf{H}_{1} $ to virtually update the model without changing the original model. 
Use $\nabla_{\psi}\mathbf{H}_{1} $ to virtually update the model without changing the original model. 

\paragraph{Step 2:} Calculate loss $\mathcal{L}_{\mathbf{H}}$ on the virtual model to compute the gradients w.r.t. the parameters in the first step:

% \begin{equation}
%     \nabla_{\phi} \mathbf{g}_{2} = \frac{\partial \mathcal{L}_{\mathbf{H}}}{\partial \phi^{(0)}} = \frac{\partial J_{\mathbf{H}}(\phi^{(0)} - \alpha \frac{\partial \mathcal{L}_{CE}}{\partial \phi^{(0)}} )}{\partial \phi^{(0)}} = J_{\mathbf{H}}'(.) +  \frac{\partial }{\partial \phi^{(0)}}(\phi^{(0)} - \alpha \frac{\partial \mathcal{L}_{CE}}{\partial \phi^{(0)} }) = J_{\mathbf{H}}'(.) - {\color{blue} \alpha \frac{\partial^2 \mathcal{L}_{CE}}{\partial {\phi^{(0)}}^2}} 
% \end{equation}

\begin{equation}
    \nabla_{\psi} \mathbf{H}_{2} = \frac{\partial \mathcal{L}_{\mathbf{H}}}{\partial \psi^{(0)}} = \frac{\partial J_{\mathbf{H}}(\psi^{(0)} - \beta \frac{\partial \mathcal{L}_{CE}}{\partial \psi^{(0)}} )}{\partial \psi^{(0)}} = J_{\mathbf{H}}'(.) +  \frac{\partial }{\partial \psi^{(0)}}(\psi^{(0)} - \alpha \frac{\partial \mathcal{L}_{CE}}{\partial \psi^{(0)} }) = J_{\mathbf{H}}'(.) - {\color{blue} \alpha \frac{\partial^2 \mathcal{L}_{CE}}{\partial {\psi^{(0)}}^2}} 
\end{equation}

% where $J_{\mathbf{H}}$ and $J_{\mathbf{H}}'$ represent the loss function used to calculate $\mathcal{L}_H$ as in Equation \ref{diffnorm}. To obtain $\nabla_{\phi} \mathbf{g}_{2}$ and $\nabla_{\psi} \mathbf{H}_{2} $, we need to compute the second-order derivatives as shown in blue. In Appendix \ref{appendix_secondorder}, we show that even a first-order approximation achieves similar performance, thus in practice, we can use a first-order approximation. 

where $J_{\mathbf{H}}$ and $J_{\mathbf{H}}'$ represent the loss function used to calculate $\mathcal{L}_H$ as in Equation \ref{diffnorm}. To obtain $\nabla_{\psi} \mathbf{H}_{2} $, we need to compute the second-order derivatives as shown in blue. In Appendix \ref{appendix_secondorder}, we show that even a first-order approximation achieves similar performance, thus in practice, we can use a first-order approximation. 

\paragraph{Step 3:} Finally, we need to compute the final gradients to update the main model. The gradients in the first step correspond to the changes that are useful for the current batch, and the gradients in the second step are used for the changes that can help to avoid forgetting the previous tasks. Therefore, we use the mean of the gradients:

\begin{align}
    % \nabla_{\phi} \mathbf{g}= \frac{\nabla_{\phi} \mathbf{g}_{1} + \nabla_{\phi} \mathbf{g}_{2}}{2} \\
    \nabla_{\psi} \mathbf{H} = \frac{\nabla_{\psi} \mathbf{H}_{1} + \nabla_{\psi} \mathbf{H}_{2}}{2} 
\end{align}

$\nabla_{\psi} \mathbf{H}$ is used to update the parameters of the hypernetwork at each training step.

\section{Experiments}
In this section, we conduct experiments to demonstrate the efficacy of partial weight generation using HNs. First, we justify the approach by showing that under certain conditions, full weight generation is unnecessary. We then show that HNs are more robust to noises in the stream. Finally, we test partial HNs and LR on the Split-CIFAR100 and Split-TinyImagenet benchmarks as two standard test beds and compare them for different freezing depths. Throughout all experiments, we utilize slim ResNet-18 \cite{lopez2017gradient} as the base model, which usually serves as a standard model for comparison in continual learning papers. For the hypernetwork, we employ a multi-layer perception with two hidden layers and separate weight generation heads for each layer of the main model. Additional details about the model architectures can be found in Appendix \ref{appendix_modeldetails}.

The main model architecture consists of a convolutional layer, four residual blocks, and a classification layer.  We adopt a block-wise freezing approach, wherein \textit{the freeze depth $k$} implies that the first $k$ blocks of the model are frozen including the initial convolutional layer. This allows us to create five distinct versions of the base model, denoted by ResNet-$k$ where $k=\{0, 1, 2, 3, 4\}$ determines the number of blocks up to which the model is frozen. For instance, ResNet-$0$ indicates no frozen layers, whereas ResNet-$4$ implies that all layers except the classification layer are frozen. In the HN versions, the weights of the classification layer are generated by the hypernetwork. We implement all strategies using the Avalanche library \cite{lomonaco2021avalanche}. The codebase of our implementation is publicly available at \href{https://github.com/HamedHemati/Partial-Hypernetworks-for-CL.git}{https://github.com/HamedHemati/Partial-Hypernetworks-for-CL.git}.

\subsection{Analyzing the Effect of Frozen Layers When Learning New Tasks}

\begin{figure}[t]
  \centering
%   \fbox{\rule{0pt}{2in} \rule{0.9\linewidth}{0pt}}
    \begin{subfigure}[t]{0.3\textwidth}
        \includegraphics[width=0.95\linewidth]{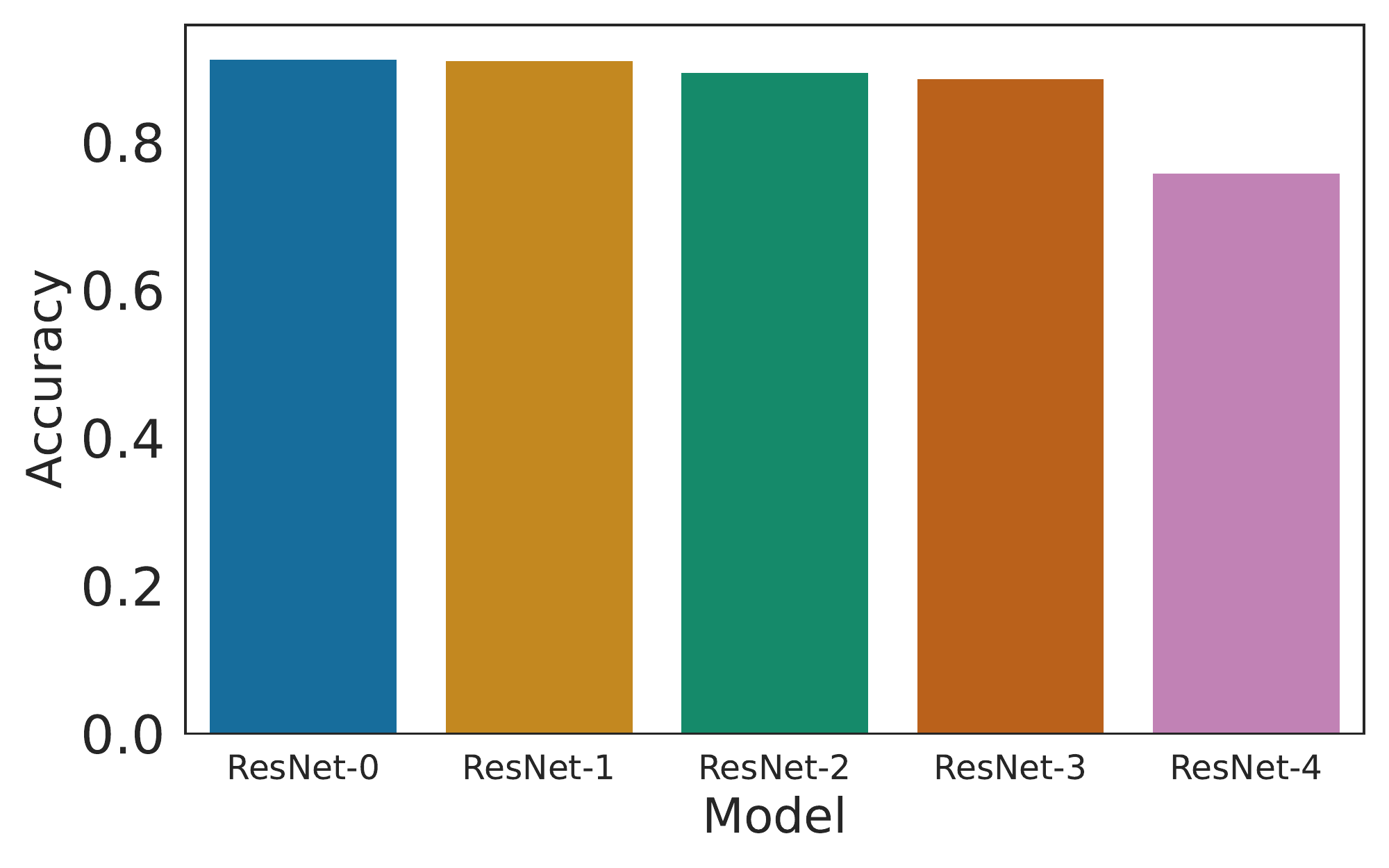}
        \caption{CIFAR-10 $\rightarrow$ CIFAR-100}
    \end{subfigure}
    \begin{subfigure}[t]{0.3\textwidth}
        \includegraphics[width=0.95\linewidth]{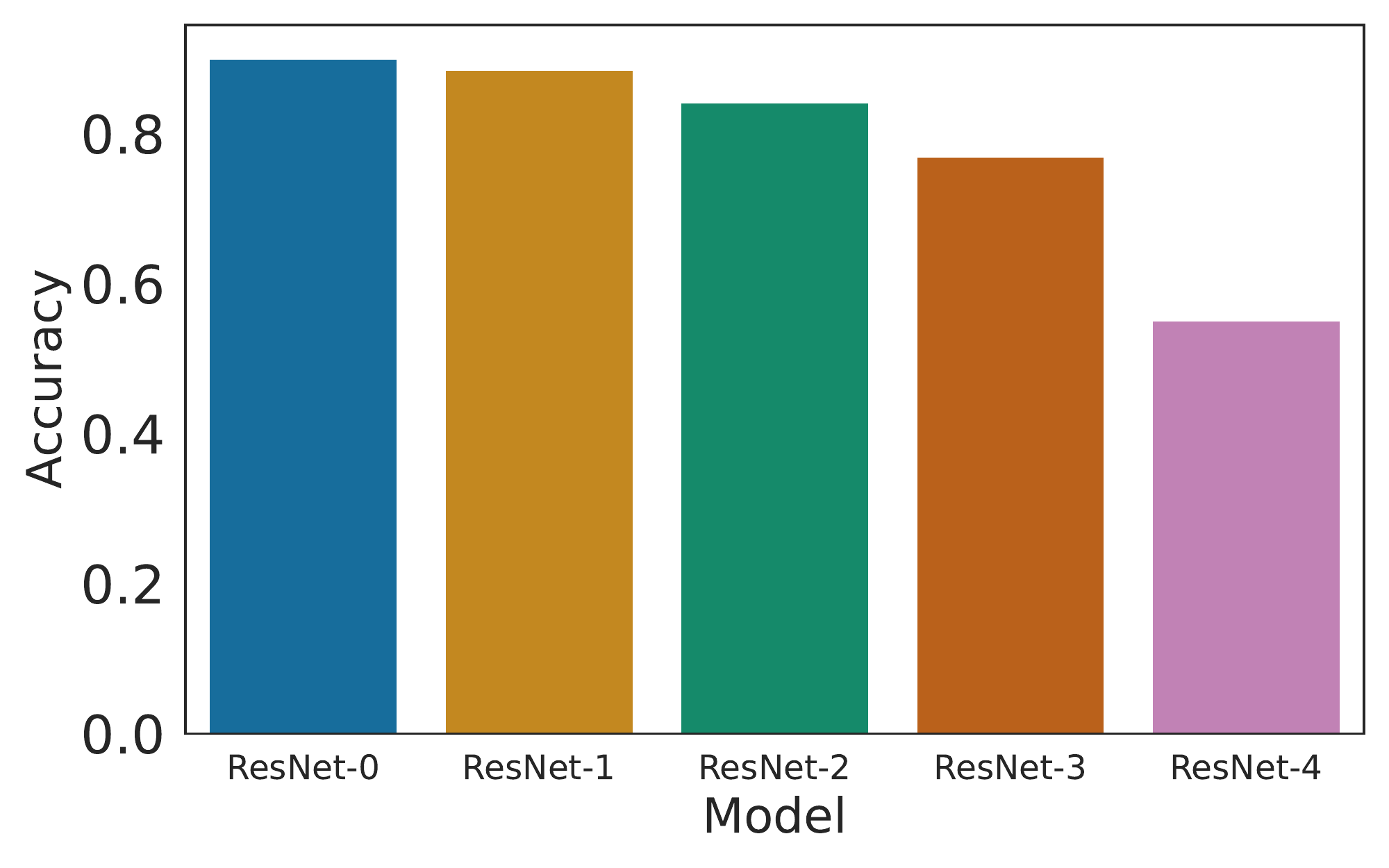}
        \caption{GTSRB $\rightarrow$ CIFAR-100}
    \end{subfigure}
    \begin{subfigure}[t]{0.3\textwidth}
        \includegraphics[width=0.95\linewidth]{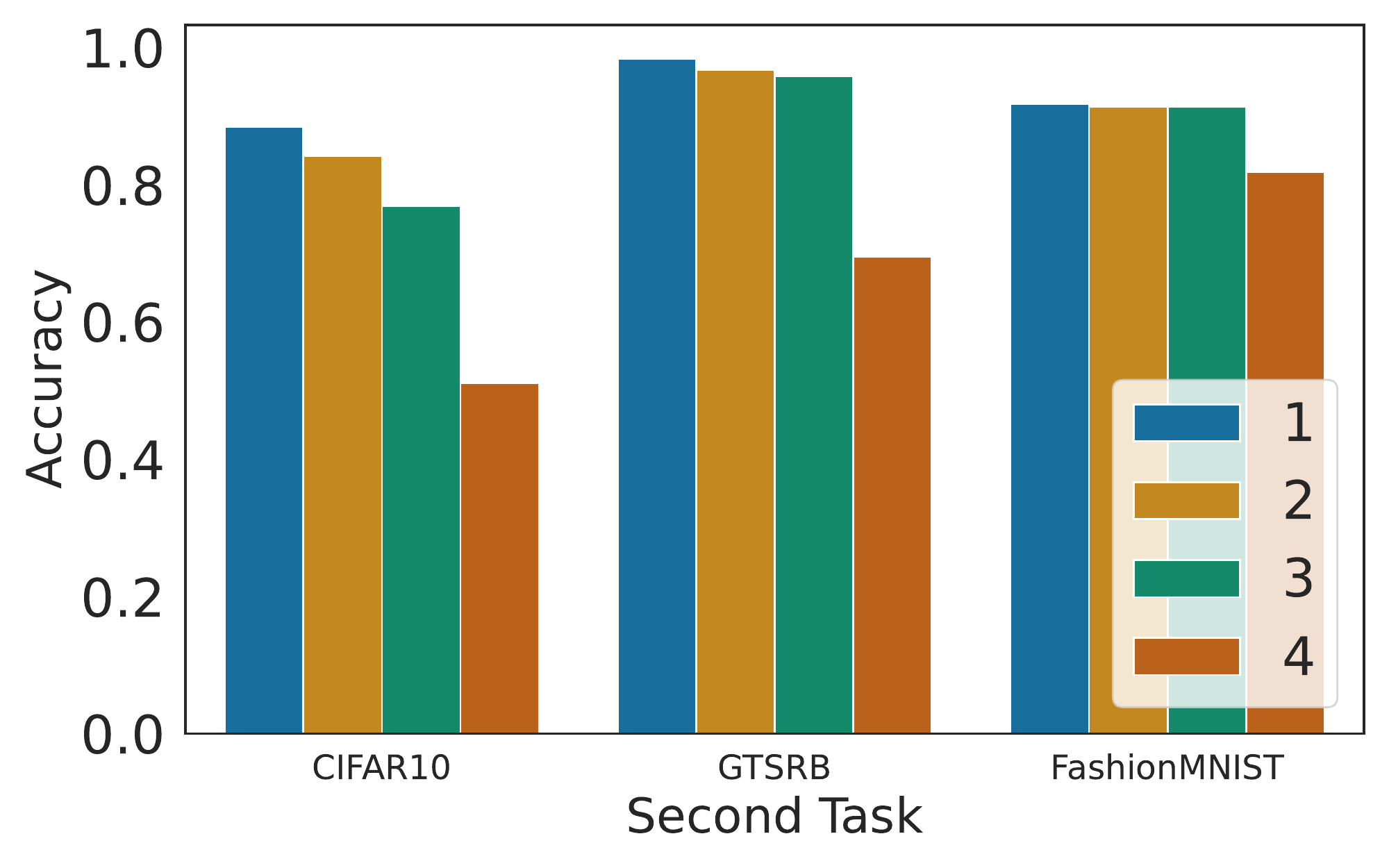}
        \caption{SVHN $\rightarrow$ \{CIFAR-10, GTSRB, FashionMNIST\}. Numbers show freeze depth.}
        % \label{fig:Product owner}
    \end{subfigure}

   \caption{Test accuracy in the second experience for different set of streams.}
   \label{fig:exp1_1}
\end{figure}

Our investigation begins by asking the following question: \\

\textit{How many layers can be frozen without a significant reduction in the performance?}

Obtaining an answer to this question provides a good starting point for applying partial weight generation via HNs. To this end, we design a small experiment using five different datasets: FashionMNIST \cite{xiao2017fashion}, SVHN \citet{netzer2011reading}, GTSRB, CIFAR-10, and CIFAR-100 \cite{krizhevsky2009learning}. By leveraging the variations and complexities across these datasets, we can assess the extent to which frozen layers negatively impact the model's performance. For this purpose, we create two sets of CL streams based on distributional similarities between experiences. In the first set, each stream entails two experiences with five random classes in each experience drawn from different distributions. For example, a stream may consist of samples from the FashionMNIST dataset in the first experience and the CIFAR-10 dataset in the second experience. We want to measure the learning accuracy in the second experience by enforcing the model to use the frozen layers learned in the first experience from a different distribution with a varying number of frozen blocks. In the second set, we similarly create streams of length two with five random classes in each experience, but we sample from CIFAR-10 for the first experience and from CIFAR-100 for the second experience. Since both datasets come from similar distributions, we want to examine how distribution similarity between experiences affects the model for changing freeze depths.

\begin{figure}[t]
  \centering
    \includegraphics[width=0.99\linewidth]{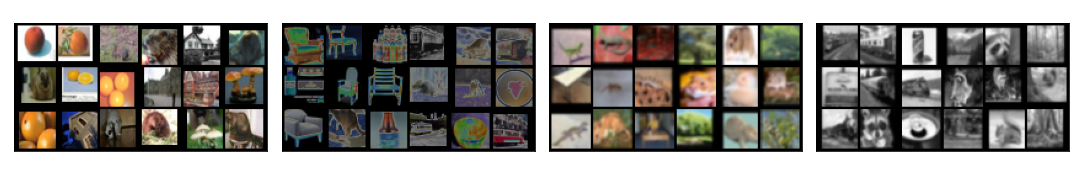}
    \caption{Random samples from  a noisy stream consisting of $4$ experiences. Each experience consists of five distinct classes. }
    \label{fig:noisy_stream}
    \vspace{-4mm}
\end{figure}

In Figure \ref{fig:exp1_1} (a and b), we show that when the model is first trained on five random classes from the CIFAR-10 dataset, learning the next experience drawn from CIFAR-100 can work well for most versions of ResNet-$k$ model as the accuracy drop is minimal. However, if we first train the model on the GTSRB dataset, learning a new experience from the CIFAR-100 dataset can be significantly affected depending on the model version, i.e., the number of frozen blocks. Moreover, in Figure \ref{fig:exp1_1} (c), we demonstrate that the complexity of subsequent experiences in a stream also plays an important role. For example, training a model on SVHN in the first experience and freezing the first three blocks doesn't cause a noticeable reduction in the performance when learning the FashionMNIST dataset since the new experience is simpler and doesn't require complex features. This suggests that the distribution variation in a stream is an important indicator in the transferability of features among experiences however, different factors such as freeze depth, the affinities between experiences, and the difficulty of the new experience can greatly impact the learning process. According to the results, we conclude that freezing more layers works only if the distributional similarity between the data from the new experience and data from the experience used to train the frozen layers is high or if the new task is simpler in terms of feature complexity. These results are aligned with the findings in \citet{yosinski2014transferable}.

\begin{wrapfigure}{r}{0.4\textwidth}
    \vspace{-15mm}
  \begin{center}
    \includegraphics[width=0.99\linewidth]{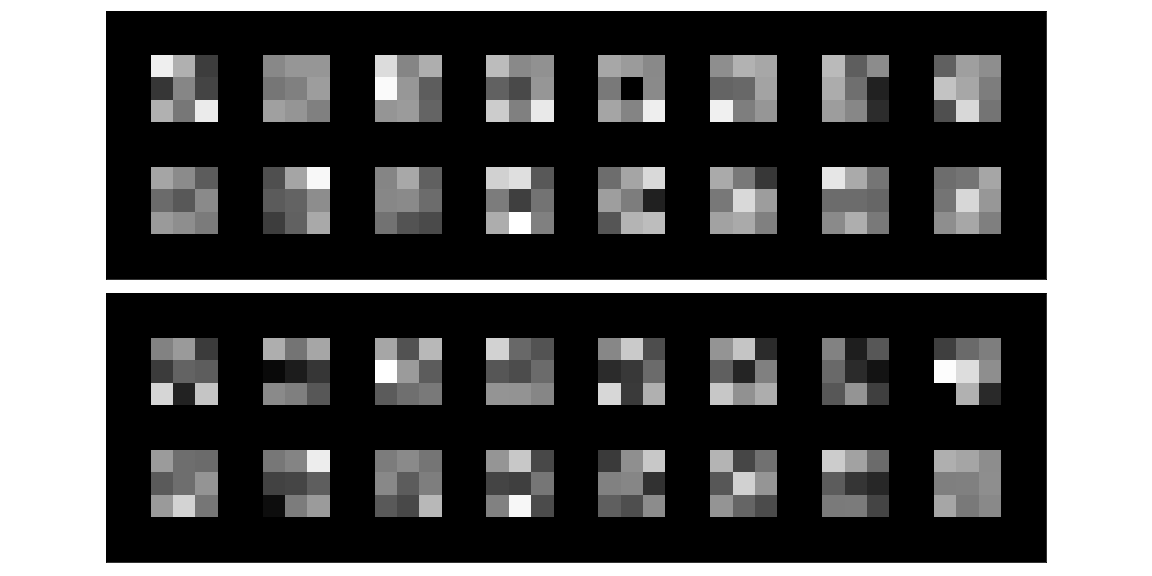}
  \vspace{-5mm}
  \end{center}
  \caption{Visualization of the filters learned for experience $2$ (above) and experience $3$ (bottom) in a partial HN with freeze-depth=$2$ in a noisy stream.}
  \label{fig:noisy_stream_filters}
  \vspace{-2mm}
  
\end{wrapfigure}

\begin{figure}[h]
    % \vspace{-4mm}
  \centering
%   \fbox{\rule{0pt}{2in} \rule{0.9\linewidth}{0pt}}
    \includegraphics[width=0.32\linewidth]{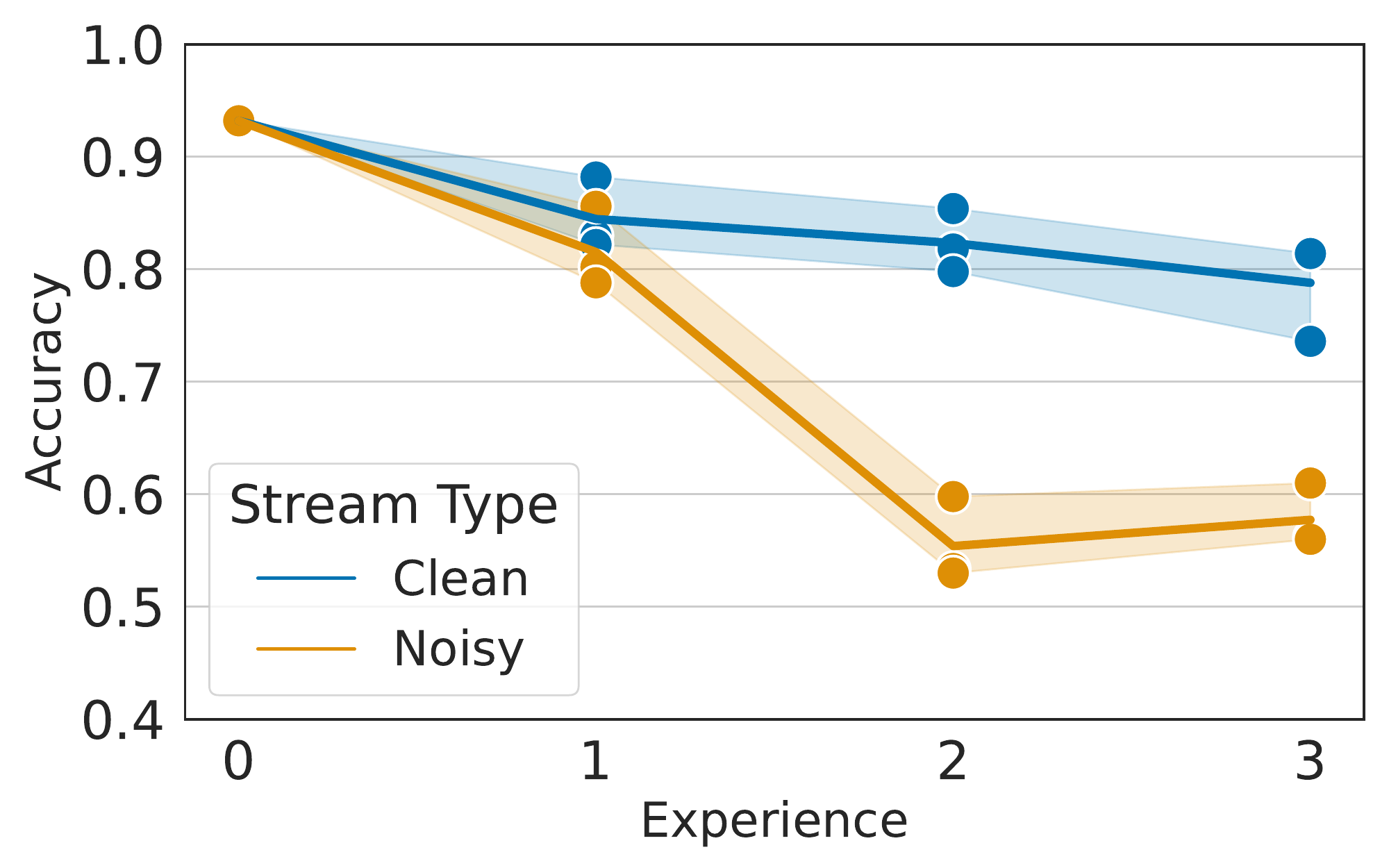}
    \includegraphics[width=0.32\linewidth]{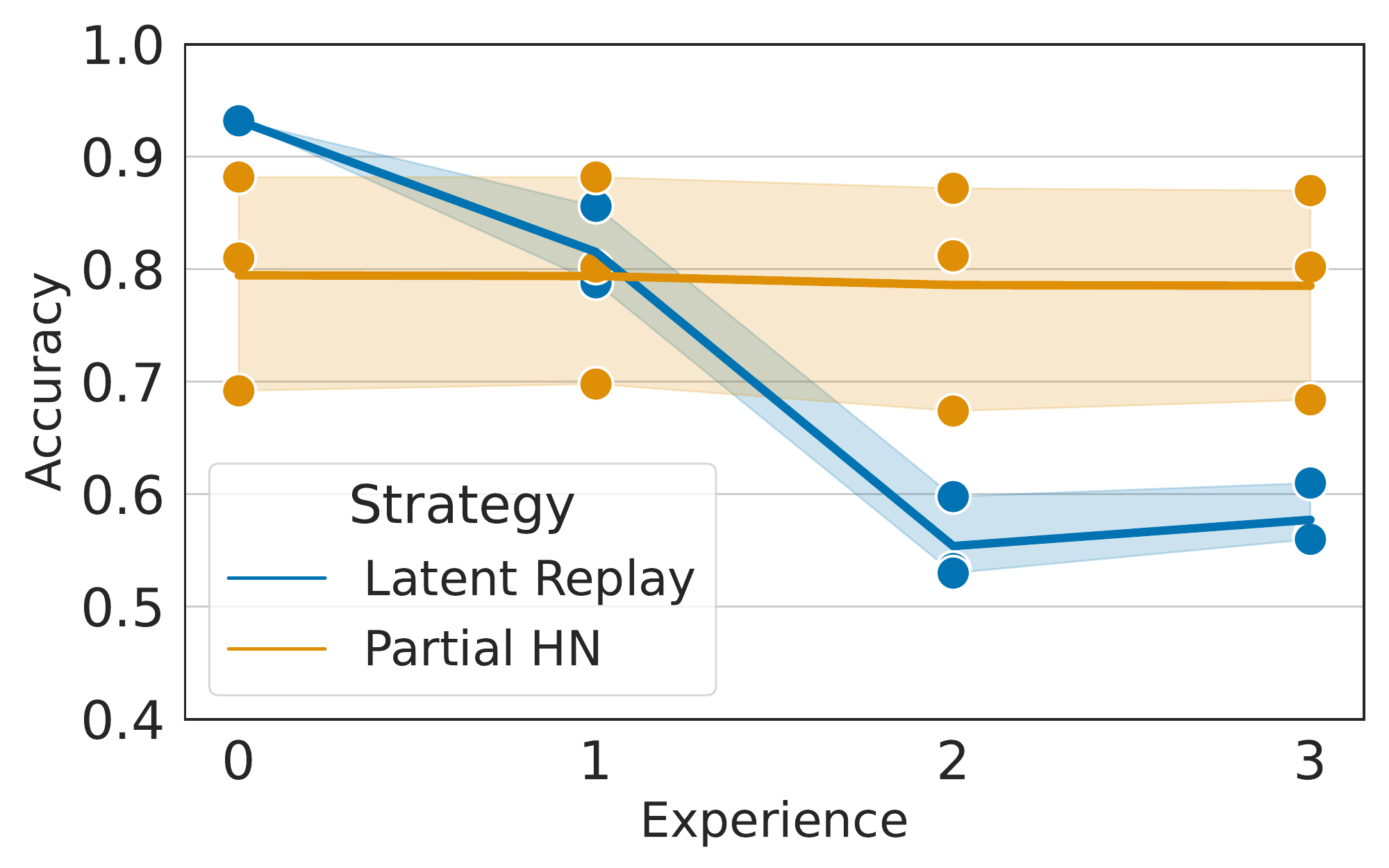}
    \includegraphics[width=0.32\linewidth]{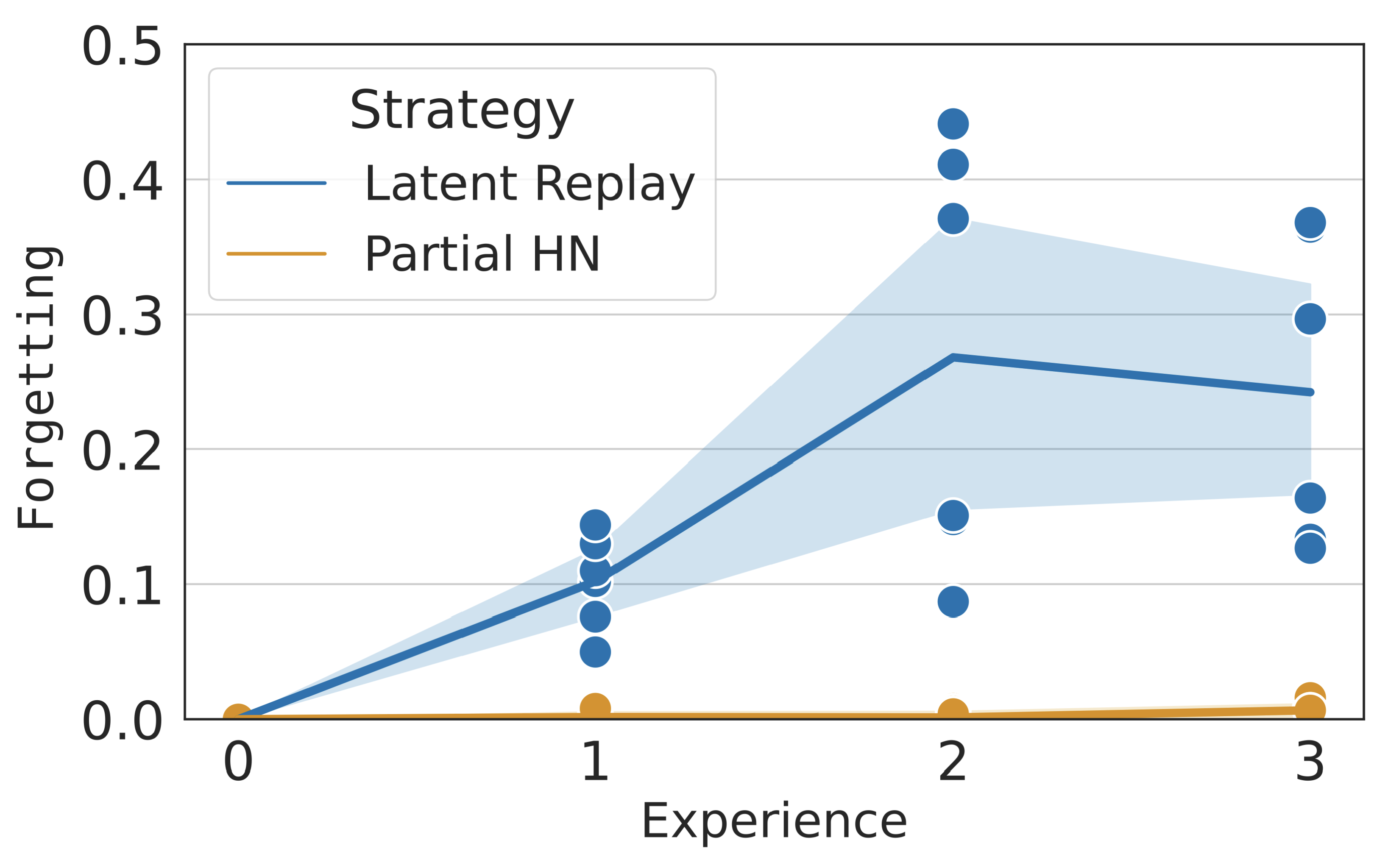}
   % \vspace{-2mm}
   \caption{Left: test accuracy of the \textbf{first experience} over time in LR with different freezing depths and replay coefficient $c=1$ for the noisy and clean versions of the same stream. Middle and Right: comparison of test accuracy and forgetting for the \textbf{first experience} between LR and Partial HN for different freezing depths in the noisy stream}.
   \label{fig:noisy_stream_results}
   \vspace{-4mm}
\end{figure}

\subsection{Robustness to Noise}

The primary objective in this experiment is to evaluate the effectiveness of HNs in a setting where latent replay methods may struggle to retain performance. We first hypothesized that in the presence of significant distribution shifts between experiences, replaying older samples with newer ones may result in a severe degradation of performance in older experiences in LR methods. Since one of the key features of HNs is their input adaptability, we sought to investigate whether having separate convolution kernels per experience can help mitigate interference between experiences in such settings.

To construct the noisy streams, we randomly shuffle CIFAR-100 classes, select the first $20$ classes, and build streams with five classes in each experience. Each experience applies a different set of  transformations to the images in its dataset. These transformations are distinct and allow us to observe the impact of significant changes in the stream over time. As illustrated in Figure \ref{fig:noisy_stream}, in our experiments we apply the following transformations: Random Solarize to experience $2$, Gaussian Blur to experience $3$, and a combination of Contrast, Gaussian Blur, and GrayScale to experience $4$.

\renewcommand{\arraystretch}{1.1}
\vspace{10pt}
\begin{wraptable}{r}{0.6\textwidth}
  \small
  \centering
  \begin{tabular}{@{}l c c c c c @{}}
    \Xhline{2\arrayrulewidth}
    \textbf{Model} & \textbf{Stream} & $\mathbf{e_1}$ & $\mathbf{e_2}$ & $\mathbf{e_3}$ & $\mathbf{e_4}$ \\
    \midrule
    % \cline{3-6}
    \multirow{2}{*}{ResNet-$2$}  & Normal & $93.2 \pm {\scriptstyle 0.2}$ & $94.3 \pm {\scriptstyle0.4}$ & $91.2 \pm {\scriptstyle 0.4}$ & $92.2 \pm {\scriptstyle 0.3}$ \\
                                 & Noisy & $93.2 \pm {\scriptstyle 0.2}$ & $91.2 \pm {\scriptstyle  0.6}$ & $89.1 \pm {\scriptstyle 1.3}$ & $91.2 \pm {\scriptstyle 0.9}$\\
    \cline{3-6}
    \multirow{2}{*}{ResNet-$2$*}  & Normal & $88.2 \pm {\scriptstyle 1.4}$ & $77.2 \pm {\scriptstyle 0.3}$ & $76.6 \pm {\scriptstyle 0.7}$ & $64.6 \pm {\scriptstyle 0.3}$ \\
                                 & Noisy & $88.2 \pm {\scriptstyle 1.4}$ & $76.6 \pm {\scriptstyle 1.3}$ & $65.0 \pm {\scriptstyle 2.1}$ & $50.0 \pm {\scriptstyle 0.8}$ \\
    % \cline{3-6}

    \midrule
    \multirow{2}{*}{ResNet-$3$}  & Normal & $93.1 \pm {\scriptstyle 0.8}$ & $94.3 \pm {\scriptstyle 0.1}$ & $89.9 \pm {\scriptstyle 0.7}$ & $92.3 \pm {\scriptstyle 0.4}$  \\
                                 & Noisy & $93.1 \pm {\scriptstyle 0.8}$ & $91.1 \pm {\scriptstyle 0.9}$ & $88.0 \pm {\scriptstyle 1.2}$ & $87.4 \pm {\scriptstyle 0.7}$  \\
    \cline{3-6}
    % \midrule
    \multirow{2}{*}{ResNet-$3$*}  & Normal & $81.0 \pm {\scriptstyle 0.6}$ & $71.4 \pm {\scriptstyle 0.5}$ & $70.3 \pm {\scriptstyle 0.4}$ & $71.5 \pm {\scriptstyle 0.4}$  \\
                                 & Noisy & $81.0 \pm {\scriptstyle 0.6}$ & $69.1 \pm {\scriptstyle 1.1}$ & $66.2 \pm {\scriptstyle 1.3}$ & $68.1 \pm {\scriptstyle 0.9}$  \\

    \midrule
    \multirow{2}{*}{ResNet-$4$}  & Normal & $93.1 \pm {\scriptstyle 0.4}$ & $91.3 \pm {\scriptstyle 0.6}$ & $85.1 \pm {\scriptstyle 0.2}$ & $89.7 \pm {\scriptstyle 0.5}$  \\
                                 & Noisy & $93.1 \pm {\scriptstyle 0.4}$ & $86.4 \pm {\scriptstyle 1.2}$ & $86.0 \pm {\scriptstyle 0.6}$ & $80.1 \pm {\scriptstyle 0.9}$  \\
    \cline{3-6}
    % \midrule
    \multirow{2}{*}{ResNet-$4$*}  & Normal & $69.1 \pm {\scriptstyle 0.3}$ & $68.6 \pm {\scriptstyle 0.2}$ & $73.4 \pm {\scriptstyle 0.2}$ & $72.1 \pm {\scriptstyle 0.1}$  \\
                                 & Noisy & $69.1 \pm {\scriptstyle 0.3}$ & $68.4 \pm {\scriptstyle 0.4}$ & $73.4 \pm {\scriptstyle 0.6}$ & $67.5 \pm {\scriptstyle 0.3}$  \\
    % \midrule
    \Xhline{2\arrayrulewidth}
  \end{tabular}
  \caption{Learning accuracy of each experience in LR and Partial HN in noisy and normal streams. Models in the partial HN strategy are marked with *. Results are obtained from three run with three random class ordering.}
  \label{tab:learningtestaccuracy}
\end{wraptable}

After training different model versions on both noisy and normal streams, we noticed that the performance of LR with replay coefficient $c=1$ is significantly harmed when the stream is noisy, especially after training on the third experience with the Gaussian Blurring noise. The issue remains despite using the same buffer size and achieving almost the same learning accuracy on the new experiences in both stream types (see Table \ref{tab:learningtestaccuracy}). However, the test accuracy of the model in partial HNs remains constant in both "normal" and "noisy" streams (Figure \ref{fig:noisy_stream_results}). That said, the learning accuracy in partial HNs significantly varies to a large extent under the same training setting depending on the number of frozen layers. As presented in Table \ref{tab:learningtestaccuracy}, all model versions in partial HNs versions have lower learning test accuracy compared to those of LR. In general, we experienced more difficulty in convergence in HN variations. Among the HN model version, ResNet-$3$ performs better on average compared to ResNet-$2,4$ in terms of learning test accuracy. Moreover, we visualize the filters of a random convolutional layer in the third block of the model in partial HNs after the third experience in Figure \ref{fig:noisy_stream_filters}. We can see that the model learns mostly different sets of filters for the new task, which can be the reason for the lower interference between experiences in partial HNs, while LR is forced to use a shared set of filters.

\subsection{Split- CIFAR-100 and TinyImageNet}

In this experiment, we test partial HNs and LR in two standard CL classification benchmarks. Both benchmarks are comprised of $20$ experiences with $5$ and $10$ distinct classes in the CIFAR and TinyImagenet benchmarks, respectively. We use a buffer size of $200$ for CIFAR-100 and $400$ for Tiny-ImageNet. The complete comparison between partial HN and LR for ResNet-$k$, $ 0 \leq k \leq 4$ is demonstrated in Table \ref{tab:table_standard_benchmarks}. It is obvious from the results that the partial HN version of each ResNet-$k$ outperforms its LR counterpart in terms of average test accuracy between experiences. More importantly, the running time in partial HNs for model versions with larger $k$'s is much lower compared to a full hypernetwork. In both benchmarks, the HN-2 strategy which uses the ResNet-2 model achieves very similar average test accuracy to the full HN version. Therefore by freezing the first two blocks in the model we can achieve the same accuracy with approximately half the training time.

\begin{figure}[t]
    % \vspace{-4mm}
  \centering
%   \fbox{\rule{0pt}{2in} \rule{0.9\linewidth}{0pt}}
    \includegraphics[width=0.32\linewidth]{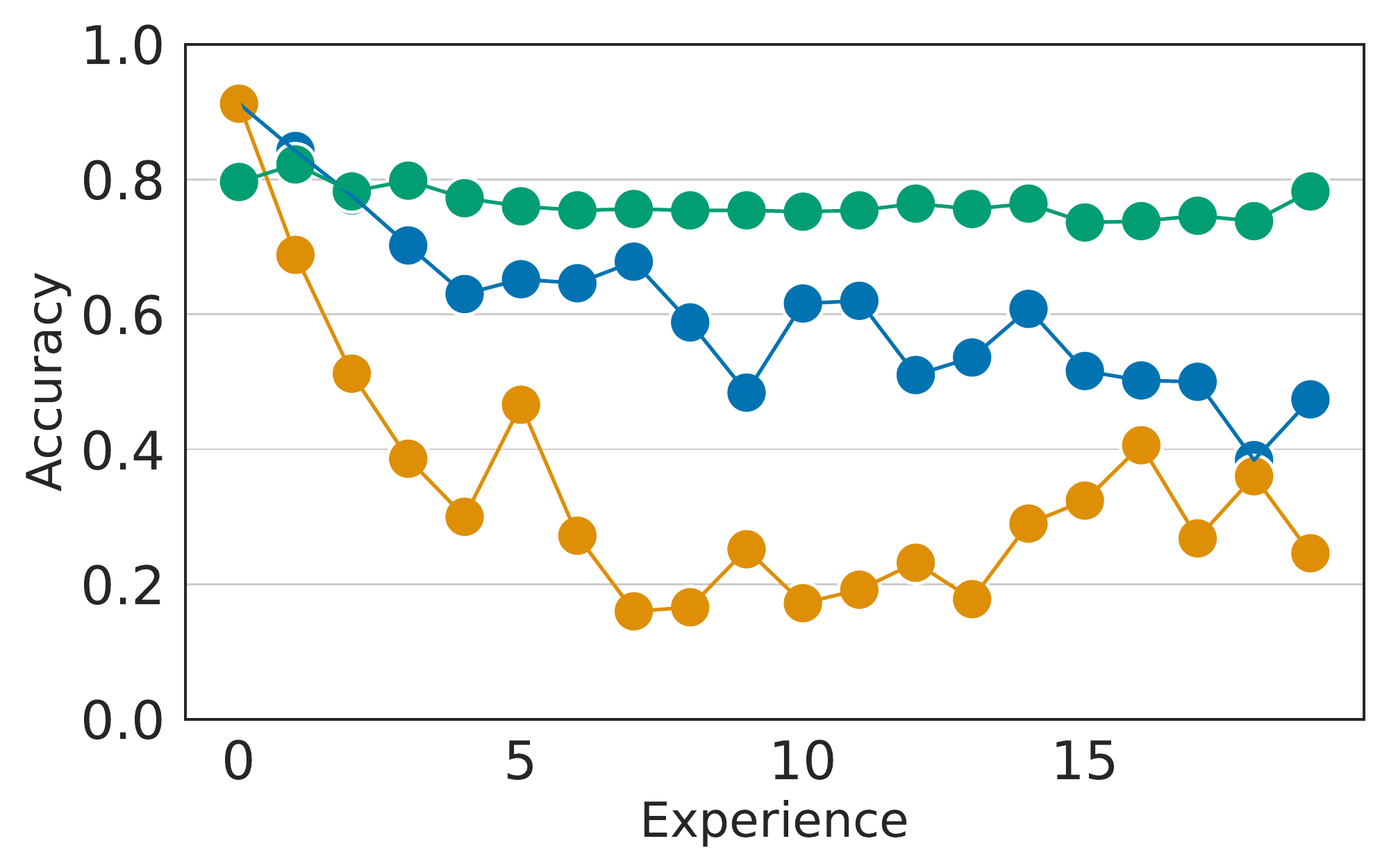}
    \includegraphics[width=0.32\linewidth]{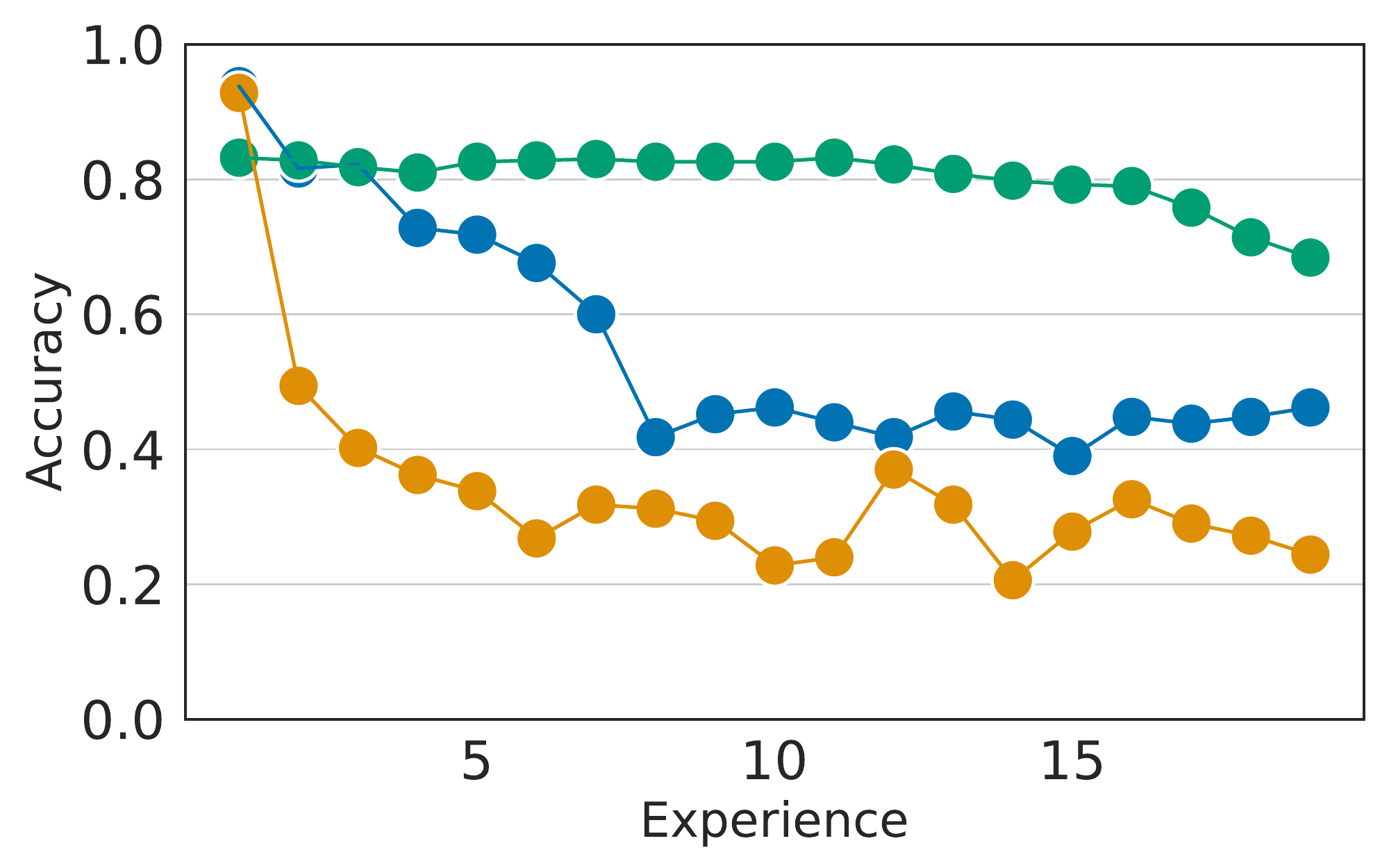}
    \includegraphics[width=0.32\linewidth]{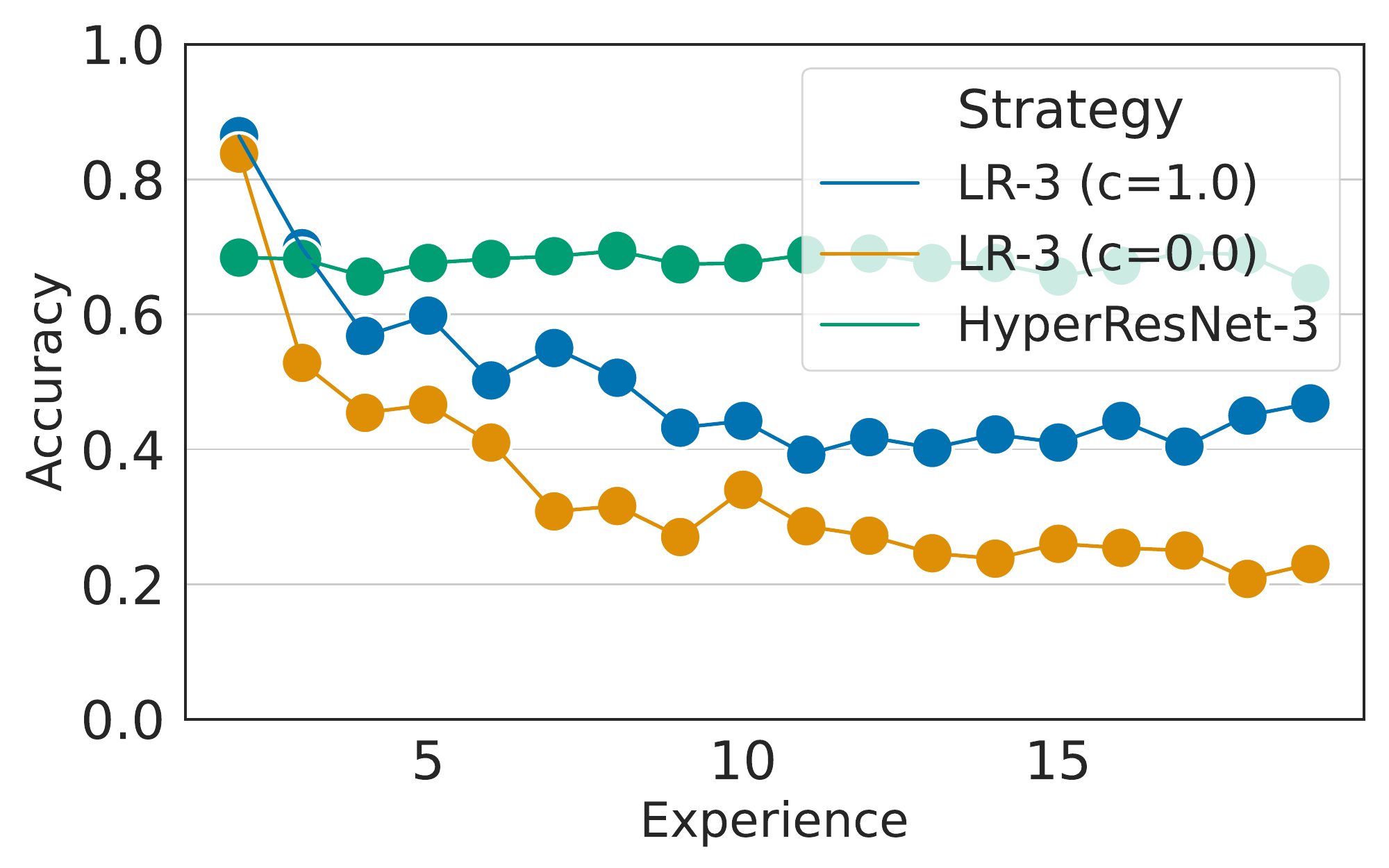}
   \caption{Test accuracy for the first (left), second(middle) and the third (right) experiences by freezing the first initial layers in ResNet-18 in the LR and partial HN methods. The coefficient $c$ in the LR method indicates the exemplar replay coefficient.}
   \label{fig:standard_benchmark_linepoint}
   \vspace{-3mm}
   % \vspace{-4mm}
\end{figure}

However, by looking at the forgetting metric, we realized that despite the low forgetting indicator, HNs achieve overall lower learning accuracy in the same training setting. This is affected by several factors such as the number of iterations, learning rate schedule, and other optimization settings. In Figure \ref{fig:standard_benchmark_linepoint} we show the test accuracy for the first three experiments over time for freezing depth $3$. The partial HN strategy is much more stable in terms of retraining accuracy compared to the LR strategies but has lower learning accuracy.

\begin{table}[ht]
\centering
\caption{Average Classification Accuracy, Forgetting and Running Time measurements for the Split-CIFAR100 and Split-Imagenet Benchmarks. The number $k$ in the LR-$k$ and HN-$k$ strategies indicates that the strategy uses ResNet-$k$ model. "Naive" refers to a strategy that performs SGD updates on the model.}
\label{tab:table_standard_benchmarks}
\renewcommand{\arraystretch}{1.2} % Adjust the spacing between rows

\begin{tabular}{c c c c c c c c c}

\Xhline{2\arrayrulewidth}
% \hline

& & \multicolumn{3}{c }{\textbf{CIFAR-100}} & &\multicolumn{3}{c}{\textbf{TinyImagenet}} \\
\textbf{Strategy} & & ACA & Forgetting & Running Time & & ACA & Forgetting & Running Time  \\
% \Xhline{2\arrayrulewidth}
% \hline

\cline{1-1} \cline{3-5} \cline{7-9}
Naive & & $33.7 \pm {\scriptstyle 0.7}$ & $58.9 \pm {\scriptstyle 2.1}$ & $\sim 120$ Min & & $20.5 \pm {\scriptstyle 1.7} $ & $48.4 \pm {\scriptstyle 0.9}$ &  $\sim 371$ Min \\

EWC & & $29.7 \pm {\scriptstyle 1.0}$ & $62.2 \pm {\scriptstyle 0.9}$ & $\sim 230$ Min & & $22.4 \pm {\scriptstyle 1.1} $ & $34.5 \pm {\scriptstyle 0.5}$ &  $\sim 431$ Min \\

% \hline
\cline{1-1} \cline{3-5} \cline{7-9}
LR-1 & & $48.1 \pm {\scriptstyle 1.2}$ & $43.0 \pm {\scriptstyle 0.8} $ & $\sim 165$ Min & & $30.6 \pm {\scriptstyle 0.9}$ & $38.1 \pm {\scriptstyle 2.1}$ & $\sim 323$ Min  \\
% \rowcolor{gray!25} % Fill this row with 25% gray color
HN-1 & & $\mathbf{61.9 \pm {\scriptstyle 1.7}}$ & $8.1 \pm {\scriptstyle 0.6}$ & $\sim 508$ Min & & $\mathbf{35.2 \pm {\scriptstyle 0.3}}$ & $08.0 \pm {\scriptstyle 1.4}$ & $\sim 905$ Min  \\
% \hline
\cline{1-1} \cline{3-5} \cline{7-9}

LR-2 & & $51.1 \pm {\scriptstyle 0.8}$ & $40.3 \pm {\scriptstyle 1.1}$ & $\sim 134$ Min & & $30.0 \pm {\scriptstyle 0.7}$ & $38.8 \pm {\scriptstyle 1.2}$ & $\sim 245$ Min  \\
% \rowcolor{gray!25} % Fill this row with 25% gray color
HN-2 & & $\mathbf{62.8 \pm {\scriptstyle 1.6}}$ & $4.1 \pm {\scriptstyle 0.5}$ & $\sim 388$ Min & & $\mathbf{39.9 \pm {\scriptstyle 0.4}}$ & $10.2 \pm {\scriptstyle 2.2}$ & $\sim 660$ Min \\
% \hline
\cline{1-1} \cline{3-5} \cline{7-9}

LR-3 & & $50.1 \pm {\scriptstyle 2.2}$ & $42.4 \pm {\scriptstyle 1.3}$ & $\sim 111$ Min & & $26.5 \pm {\scriptstyle 0.6}$ & $36.9 \pm {\scriptstyle 0.9}$ & $\sim 204$ Min  \\
% \rowcolor{gray!25} % Fill this row with 25% gray color
HN-3 & & $\mathbf{58.8 \pm {\scriptstyle 1.0}}$ & $7.4 \pm {\scriptstyle 0.9}$ & $\sim 243$ Min & & $\mathbf{31.4 \pm 1.8}$ & $5.2 \pm {\scriptstyle 0.7}$ & $\sim 416$ Min  \\
% \hline
\cline{1-1} \cline{3-5} \cline{7-9}

LR-4 & & $51.4 \pm {\scriptstyle 1.3}$ & $27.3 \pm {\scriptstyle 2.4}$ & $\sim 89$ Min & & $31.8 \pm {\scriptstyle 1.1}$ & $25.9 \pm {\scriptstyle 2.0}$ & $\sim 143$ Min  \\
% \rowcolor{gray!25} % Fill this row with 25% gray color
HN-4 & & $\mathbf{73.6 \pm {\scriptstyle 0.9}}$ & $3.1 \pm {\scriptstyle 0.4}$ & $\sim 114$ Min & & $\mathbf{36.1 \pm {\scriptstyle 1.1}}$ & $10.0 \pm {\scriptstyle 1.1}$ & $\sim 176$ Min \\
% \hline
\cline{1-1} \cline{3-5} \cline{7-9}

% \rowcolor{gray!25} % Fill this row with 25% gray color
HN-F & & $64.2 \pm {\scriptstyle 1.2}$ & $5.5 \pm {\scriptstyle 1.4}$ & $\sim 636$ Min & & $39.9 \pm {\scriptstyle 1.7}$ & $7.2 \pm {\scriptstyle 1.8}$ & $\sim 1200$ Min \\

\Xhline{2\arrayrulewidth}
\hline
\end{tabular}
\vspace{-4mm}
\end{table}

\section{Related Work}

\paragraph{Continual Learning} methods are mainly categorized into five groups \cite{wang2023comprehensive}: regularization, replay and optimization, representation and architecture. In regularization methods, the goal is to control catastrophic forgetting by adding a regularization term to the loss function that penalizes changes to the model parameters that would cause a large shift in the outputs of the previously learned tasks \cite{kirkpatrick2017overcoming, zenke2017continual, li2017learning}. In rehearsal methods, the goal is to mitigate forgetting by storing a subset of the original data for each task and replaying it during training on subsequent tasks, either by interleaving it with new data or by training the model on the replayed data after training on the new data \cite{rolnick2019experience, chaudhry2019tiny, rebuffi2017icarl, pellegrini2020latent}. Optimization-based methods attempt to alleviate forgetting by manipulating the optimization process either via projecting gradients or customized optimizer \cite{lopez2017gradient, farajtabar2020orthogonal}. In representation-based methods, the main ideas is to learn representations that can boost continual learning for example by learning more sparse representations as in \cite{javed2019meta} or by learning richer representations for transfer learning via pre-training and self-supervised learning \cite{gallardo2021self, mehta2021empirical}. In the architecture-based methods, the focus is on the role of the model's architecture in CL with the assumption that model needs to learn task-specific parameters which is itself split to sub-groups: parameter isolation and dynamic architecture. In parameter isolation, the model architecture is fixed and different subsets of parameters are assigned to different tasks, for example by masking parameters \cite{serra2018overcoming, mallya2018piggyback} or model decomposition such as \cite{miao2022continual, yoon2019scalable, kanakis2020reparameterizing}. In dynamic architecture methods, new modules are added over time for solving newer tasks such as \cite{rusu2016progressive, aljundi2017expert, xu2018reinforced}. Our method is can be categorized into the parameter isolation sub-group in the architecture group.

\paragraph{Hypernetworks} are a type of neural network architecture that have gained significant popularity in recent years. The weight generation mechanism in HNs enables them to be applied to several applications such as neural architecture search \cite{zhang2018graph} and conditional neural networks \cite{galanti2020modularity}.  Although the origin of hypernetworks dates back to the work of Hinton and Salakhutdinov in the 1990s \cite{schmidhuber1992learning}, it was not until recently that they became a popular topic of research in deep learning \cite{ha2016hypernetworks}. In the context of CL, the work from \cite{von2019continual} proposed to use hypernetwork to generate the full weights of a neural network. Morever, \cite{ehret2020continual} proposed to use hypernetwork for more effective continual learning compared to the weight regularization-based methods. A probabilistic version of hypernetworks for CL for task-conditioned CL was proposed by \cite{henning2021posterior}.

\paragraph{Latent Replay}  methods propose to store or generate activations from intermediate layers of a neural network, which are then replayed at the same intermediate layers to prevent forgetting when new tasks are learned during the CL process.  Intuitively, Latent Replay methods assume that low-level representations can be better shared across tasks and, hence, require less adaptation. To this end, low layers are either regularized \cite{liu2020generative}, trained at a slower pace \cite{pellegrini2020latent}, or frozen, as in \cite{van2020brain, hayes2019remind}. Freezing the low layers significantly reduces the computational cost of CL by reducing the number of trainable parameters. These methods rely on the assumption that feature extractors can produce meaningful features and should be preferred in applications with limited compute. In \cite{ostapenko2022foundational}, the efficacy of pre-trained vision models for downstream continual learning is studied, exploring the trade-off between CL in raw data and latent spaces, comparing pre-trained models in benchmarking scenarios, and identifying research directions to increase efficacy.

\paragraph{Transfer Learning} is a highly popular paradigm in deep learning that enables sample-efficient learning by partially or completely fine-tuning a pre-trained model for a novel task. Many researchers have studied transfer learning to understand \textit{when and how} transfer learning works. Yosinski et al. \cite{yosinski2014transferable} quantify the generality vs. specificity of deep convolutional neural networks, showing transferability issues and boosting generalization by transferring features. They demonstrate that the first layer of AlexNet can be frozen when transferring between natural and man-made subsets of ImageNet without performance impairment, but freezing later layers produces a substantial drop in accuracy. Neyshabur et al. \cite{neyshabur2020being} analyze the transfer of knowledge in deep learning and show that successful transfer results from learning low-level statistics of data. They also show that models trained from pre-trained weights stay in the same basin in the loss landscape. Kornblith et al. \cite{kornblith2019similarity} examine methods for comparing neural network representations, show the limitations of canonical correlation analysis (CCA), and introduce a similarity index based on centered kernel alignment (CKA) that can identify correspondences between representations in networks trained from different initializations. The work by Kornblith et al. \cite{kornblith2019better} evaluates the relationship between architecture and transfer learning by comparing various networks in different classification benchmarks. They find that pre-training on ImageNet provides minimal benefits for fine-grained image classification tasks. Moreover, He et al. \cite{he2019rethinking} show that transfer even between similar tasks does not necessarily improve performance, and random initialization can achieve a better solution given enough iterations.

\section{Conclusion and Discussion}
This study explores the application of hypernetworks for partial weight generation in the context of continual learning. The research question that guided our investigation was \textit{whether it is necessary to generate all the weights of a neural network when using hypernetworks} to learn from a stream of experiences. Our approach to answering this question was inspired by recent work in latent replay for CL, and we hypothesized that generating only a subset of the model parameters via hypernetworks could be a practical method for preserving previous knowledge while adapting to new information.

To test this hypothesis, we proposed decomposing the model into two parts and generating the weights of the second part of the main neural network. By generating only a subset of the model parameters in the final layers of the neural network, we showed that we could speed up the learning process and thus reduce the computational costs, and preserve previous knowledge, compared to the full-weight generation approach.  Furthermore, we studied the impact of different decomposition depths in various settings on the final performance of both latent replay and hypernetwork variants.  To further justify the use of partial hypernetworks, we also showed through our experiments that hypernetworks are more robust to strong noise in the stream in terms of negative interference between tasks compared to latent replay. The low interference in hypernetworks suggests that partially generating weights can effectively preserve previous knowledge when data from noisy distribution are replayed with clean data. Our findings indicate that partial weight generation via hypernetworks is a promising approach for preserving previous knowledge while adapting to new information.

In spite of their ability to preserve knowledge effectively, our findings indicate that partial HNs have lower learning accuracy when trained under the same training conditions. Additionally, our research only focused on using hypernetworks for weight generation in the second part of the model. A promising avenue for future inquiry would be to explore methods for enhancing learning accuracy while reducing training time in HN-based CL strategies. Alternatively, we could investigate the feasibility of using adaptive layers between pre-trained models to generate model weights partially.

In summary, this research contributes to our understanding of the effectiveness of partial hypernetworks in continual learning and provides practical guidelines for optimizing them. We believe that our work will facilitate the development of more computationally efficient algorithms for continual learning using hypernetworks. We encourage further investigation into the relationship between hypernetworks and continual learning, and we hope that our research will inspire future work in this area.

\bibliography{collas2023_conference}
\bibliographystyle{collas2023_conference}

\newpage

\appendix
\section{Model Details} \label{appendix_modeldetails}
In this section, we elaborate on the architecture details of both the hypernetwork and the main model.

\begin{figure}[h]
\centering
\includegraphics[width=0.90\linewidth]{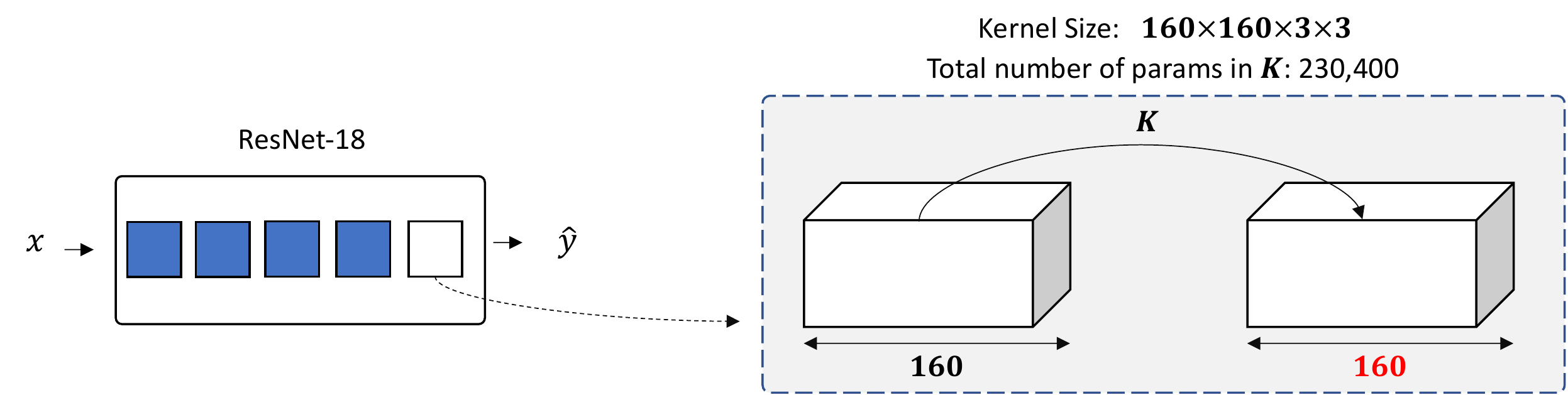}
\caption{Number of parameters in a convolution module in the final residual layer of the main model.}
\label{fig:num_params}
\end{figure}

\subsection{Architecture}

We employ a slimmed version of ResNet-18 as our main model, which comprises a single convolutional layer, four residual blocks, and a linear classifier. The model consists mainly of convolution, batch-norm, and linear module types. For the LR strategy, we utilized a multi-head classifier in the final layer, which we implemented using the \texttt{MultiHeadClassifier} module from Avalanche \footnote{\url{https://github.com/ContinualAI/avalanche/blob/master/avalanche/models/dynamic_modules.py}}. In the HN versions, the weights of the linear classifier are generated for each experience separately and we set the number of output neurons equal to the maximum number of classes in each experience.

For the hypernetwork, we use an MLP with two hidden layers. The input to the hypernetwork is the task ID. The naive way to generate weights of a model would be to generate all weights with a linear mapping directly after the second hidden layer. However, this method can become very expensive if the total number of parameters in the main model is high. For example, if the main model has $300,000$ parameters, and the second hidden layer of the hypernetwork has $100$ units, the number of parameters in the final linear layer would be $300,000 \times 100 = 30,000,000$ parameters. Instead, like most previous methods \citet{von2019continual, henning2021posterior} we propose a different way to generate the weights of the main model according to the structure of the architecture. In our setting, the final layer of the MLP outputs the weights of the main model for each module separately, thus, the total number of heads in the HN is equal to the total number of modules in the main model. We use three types of HN heads convolutional, batch-norm, and linear. Since most parameters are inside the final convolutional modules, we propose a simple yet efficient channel-wise weight generation for the convolutional layers that can be used for any convolutional neural network.

\begin{figure}[h]
\centering
\includegraphics[width=0.95\linewidth]{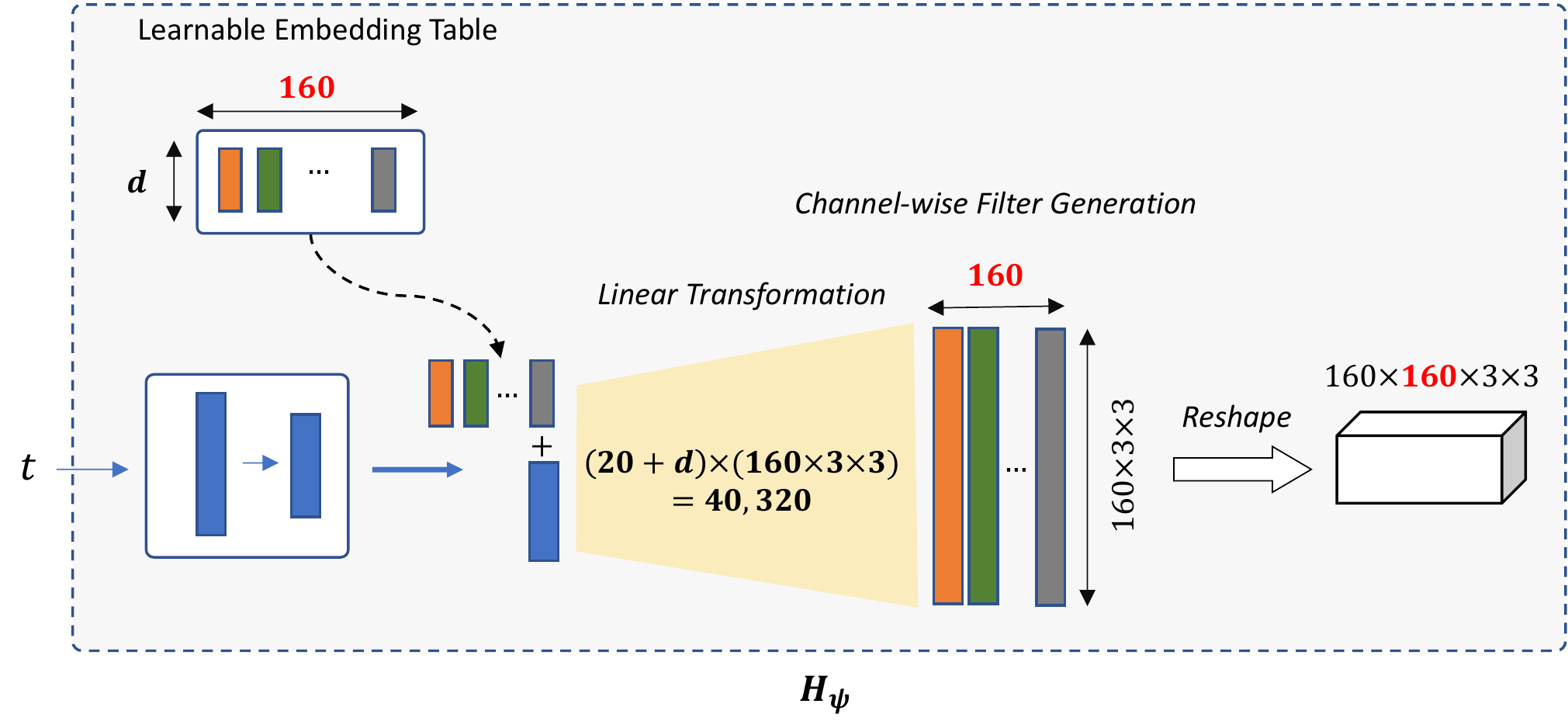}  
\caption{Channel-wise weight generation for a convolutional layer with $160$ input and output channels and a kernel size of $3 \times 3$.}
\label{fig:hnet_filters}
\end{figure}

\subsection{Channel-wise Weight Generation}
The number of parameters for a convolution module with $a$ input channels, $b$ output channels, and a kernel size of $k$ equals $a \times b \times k \times k$. If the number of units in the second hidden layer is $h$, then by direct weight generation, we will have $h \times a \times b \times k \times k$ parameters. Making $h$ smaller may work up to a certain level, but if $h$ is too small the HN will not be able to generate good filters. Therefore, we need to solve this issue by handling $a$ or $b$. We use a look-up table for $a$ to reduce its dimensionality. By assigning a learnable look-up vector with dimension $d < a$, we can condition the weight generation for the kernel of each output channel with $(h+d) \times b \times k \times k$. We show this process in Figure \ref{fig:hnet_filters}. Usually, very small $d$'s (e.g. $d=8$) work well. Hence the final number of parameters in the HN head will be $d \times a$ for the learnable lookup table plus $(h+d) \times b \times k \times k$ which is in most cases much smaller than the original number of parameters. Moreover, this type of weight generation also considers the structure of convolutional layers. We compare levels of compression in the final by setting different values to $d$ in Table \ref{tab:compression}.

\begin{table}[ht]
\centering
\caption{Compression level for different $d$'s. Number of hidden units in the second hidden layer is equal to $32$.}
\label{tab:compression}
\renewcommand{\arraystretch}{1.2} % Adjust the spacing between rows

\begin{tabular}{c c c c c}

\Xhline{2\arrayrulewidth}
% \hline
$\mathbf{d}$ & &\textbf{Number of Parameters} & & \textbf{Compression Level} \\
\cline{1-1} \cline{3-3} \cline{5-5} 
$4$  & &  $451661$ & &$59$\% \\
$8$  & &  $496489$ & &$55$\% \\
$16$  & & $586145$ & & $47$\% \\
$32$  & & $765457$ & & $31$\% \\
$64$  & & $1124081$ & & $0$\% \\

\Xhline{2\arrayrulewidth}

\end{tabular}
\end{table}

\section{Ablation Study}

In this section, we conduct an ablation study for the look-ahead mechanism and the computation of the second-order derivatives.

\subsection{Look-Ahead vs. No Look-Ahead} \label{appendix_lookahead}
The look-ahead mechanism enables a more accurate calculation of the gradients for the hypernetwork. Here, we show that applying look-ahead-based updates can be beneficial, especially for ResNet-$4$, without significantly increasing computation time. In Table \ref{tab:look_ahead}, we compare partial HNs with and without look-ahead updates for different versions of ResNet-$k$ for the same random seed and class order in the stream. The results show that in most cases, the look-ahead update improves the ACA at the end of the stream. Furthermore, we visualize the test accuracy for experiences $1$ to $8$ in a random run for both update modes in Figure \ref{fig:lookahead}. The visualizations indicate that except for the second, in the rest of the experiences, the strategy with look-ahead update performs overall better.

\begin{figure}[h]
    % \vspace{-4mm}
  \centering
%   \fbox{\rule{0pt}{2in} \rule{0.9\linewidth}{0pt}}
    \includegraphics[width=0.24\linewidth]{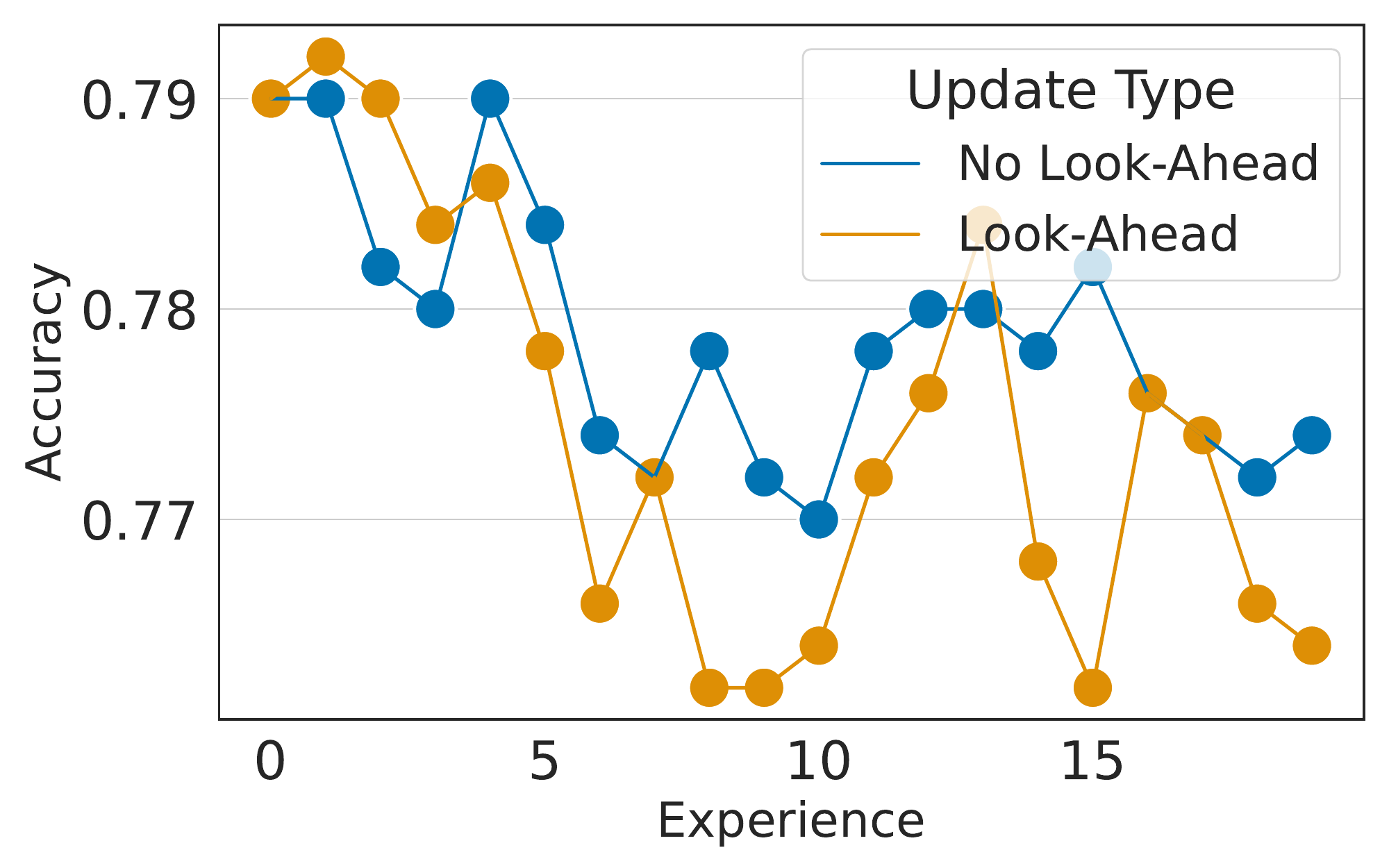}
    \includegraphics[width=0.24\linewidth]{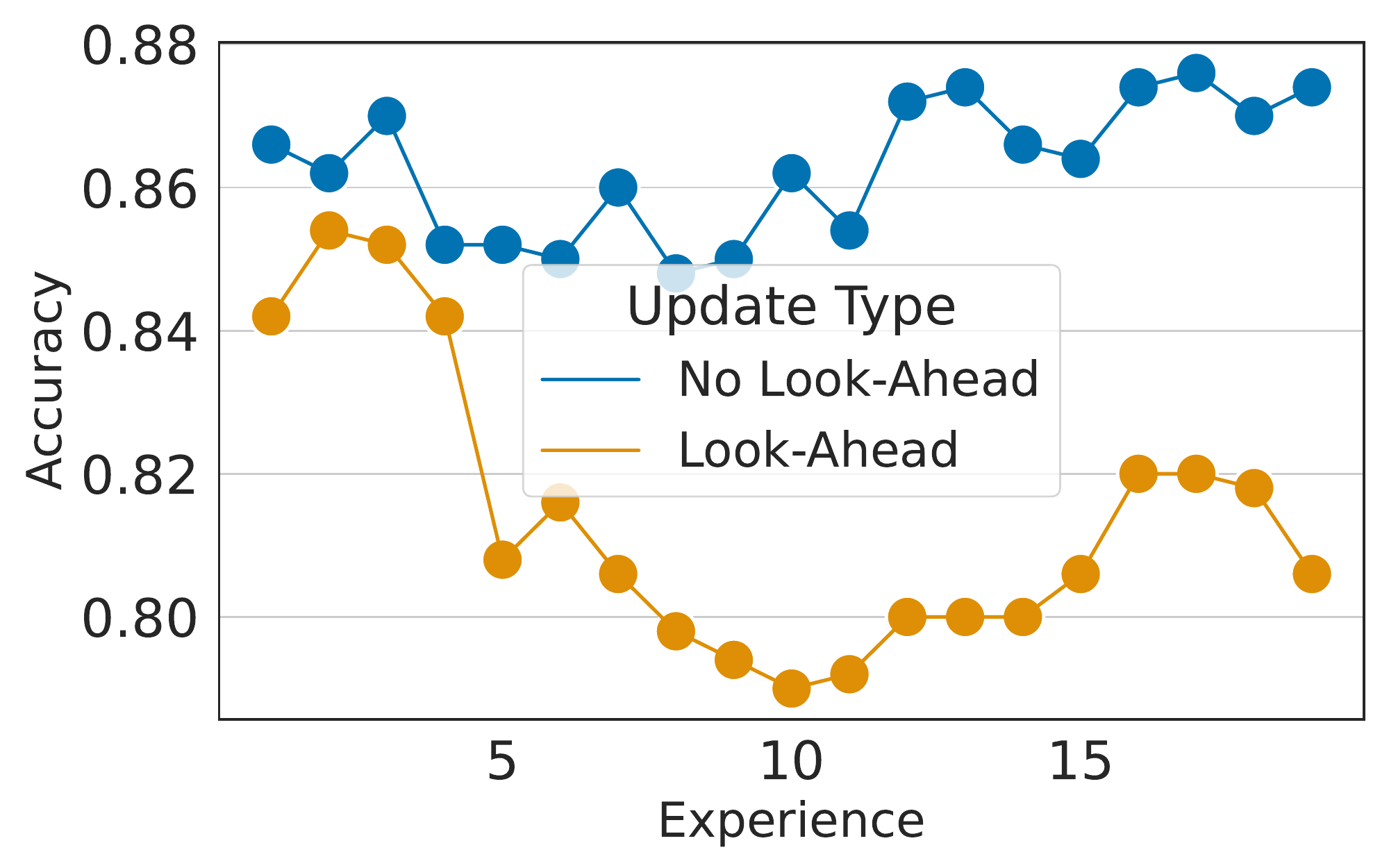}
    \includegraphics[width=0.24\linewidth]{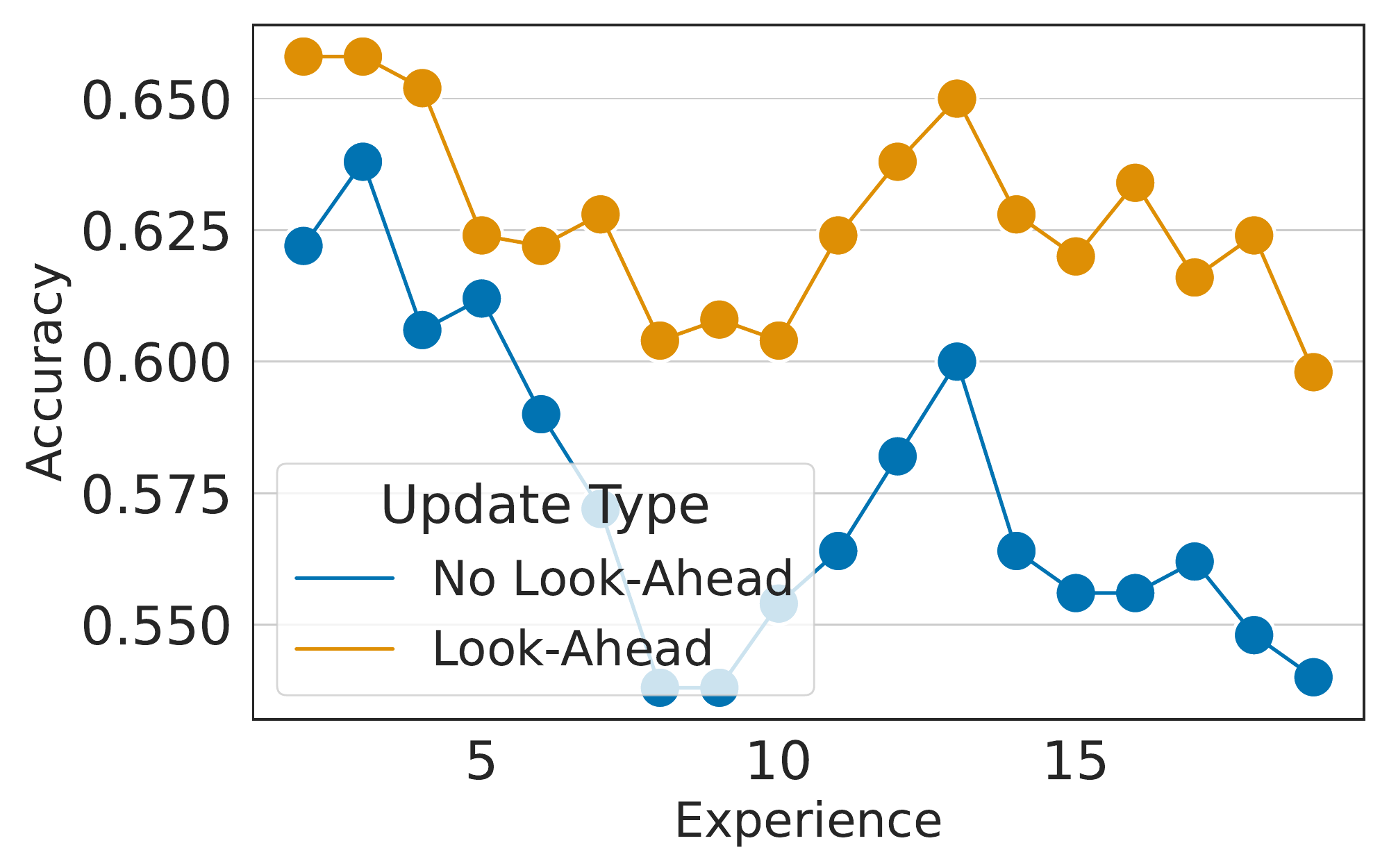}
    \includegraphics[width=0.24\linewidth]{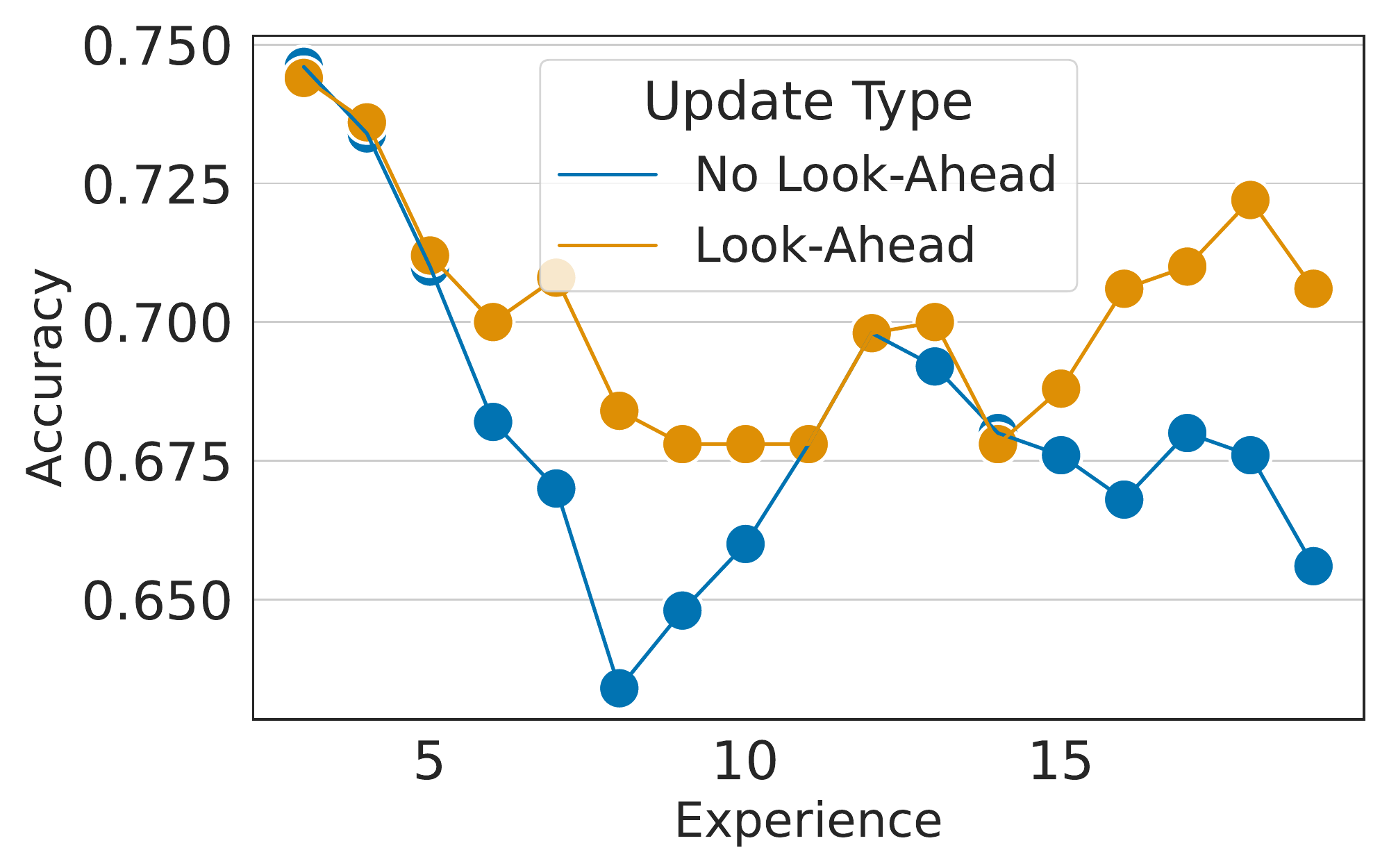}
    \includegraphics[width=0.24\linewidth]{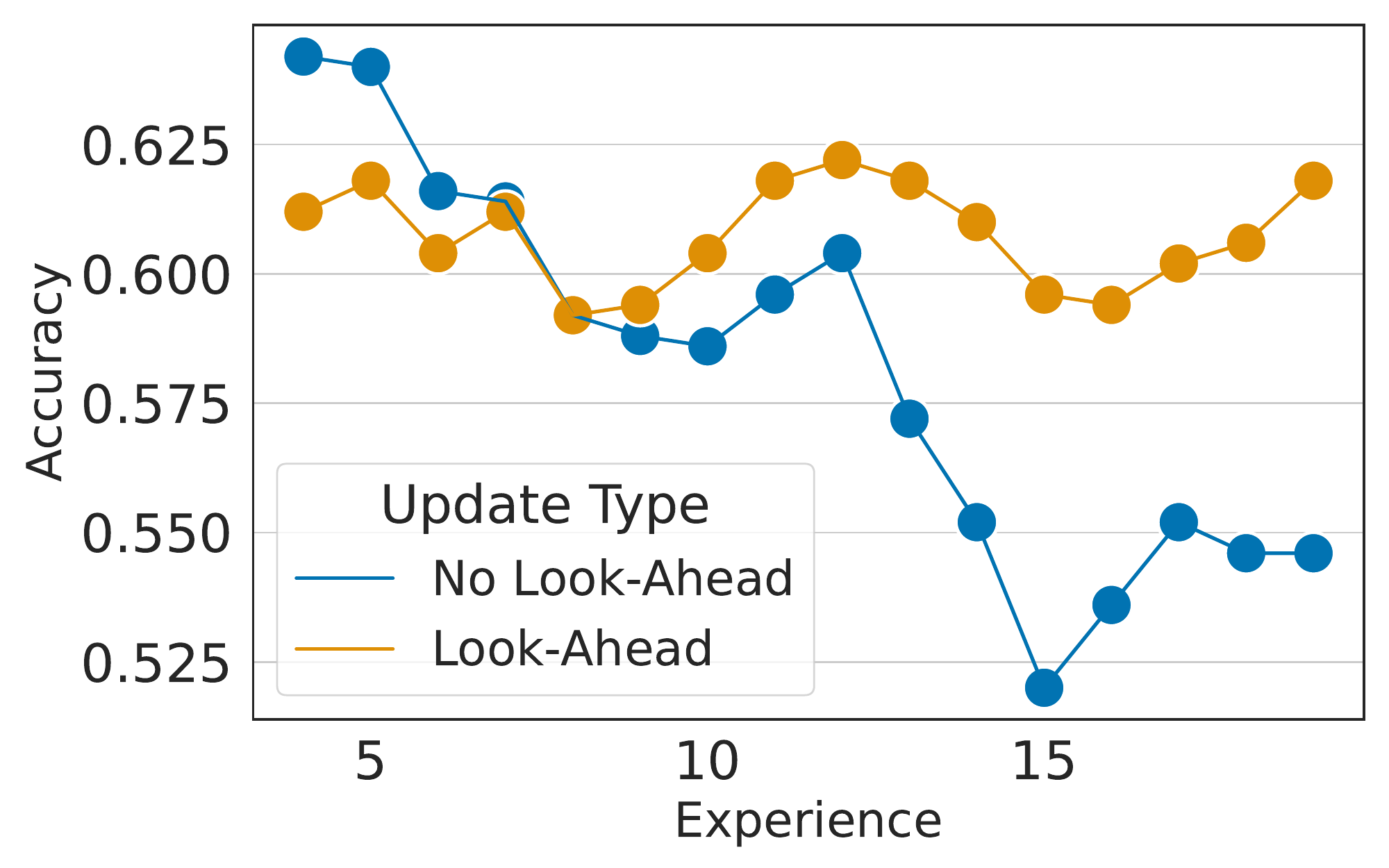}
    \includegraphics[width=0.24\linewidth]{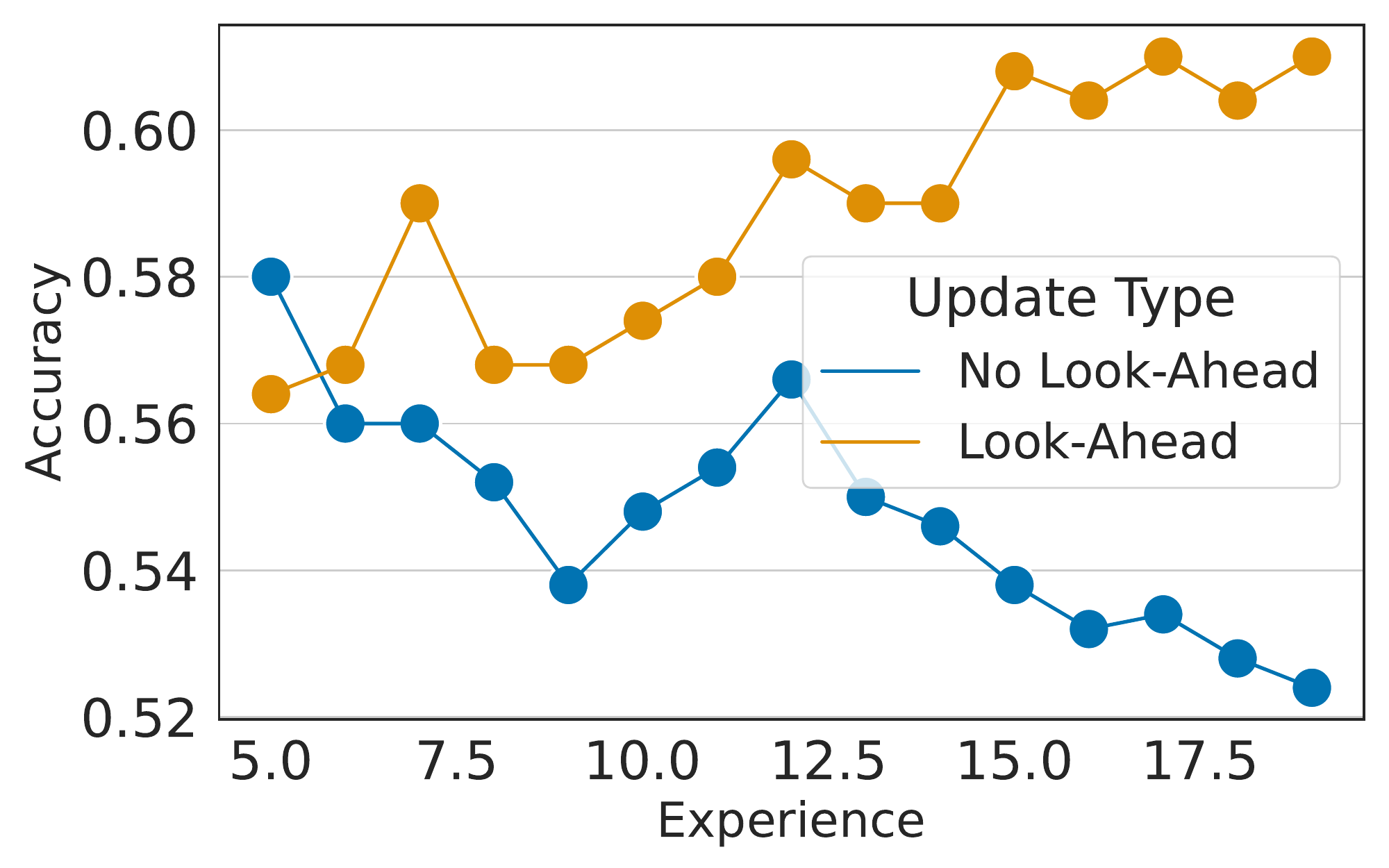}
    \includegraphics[width=0.24\linewidth]{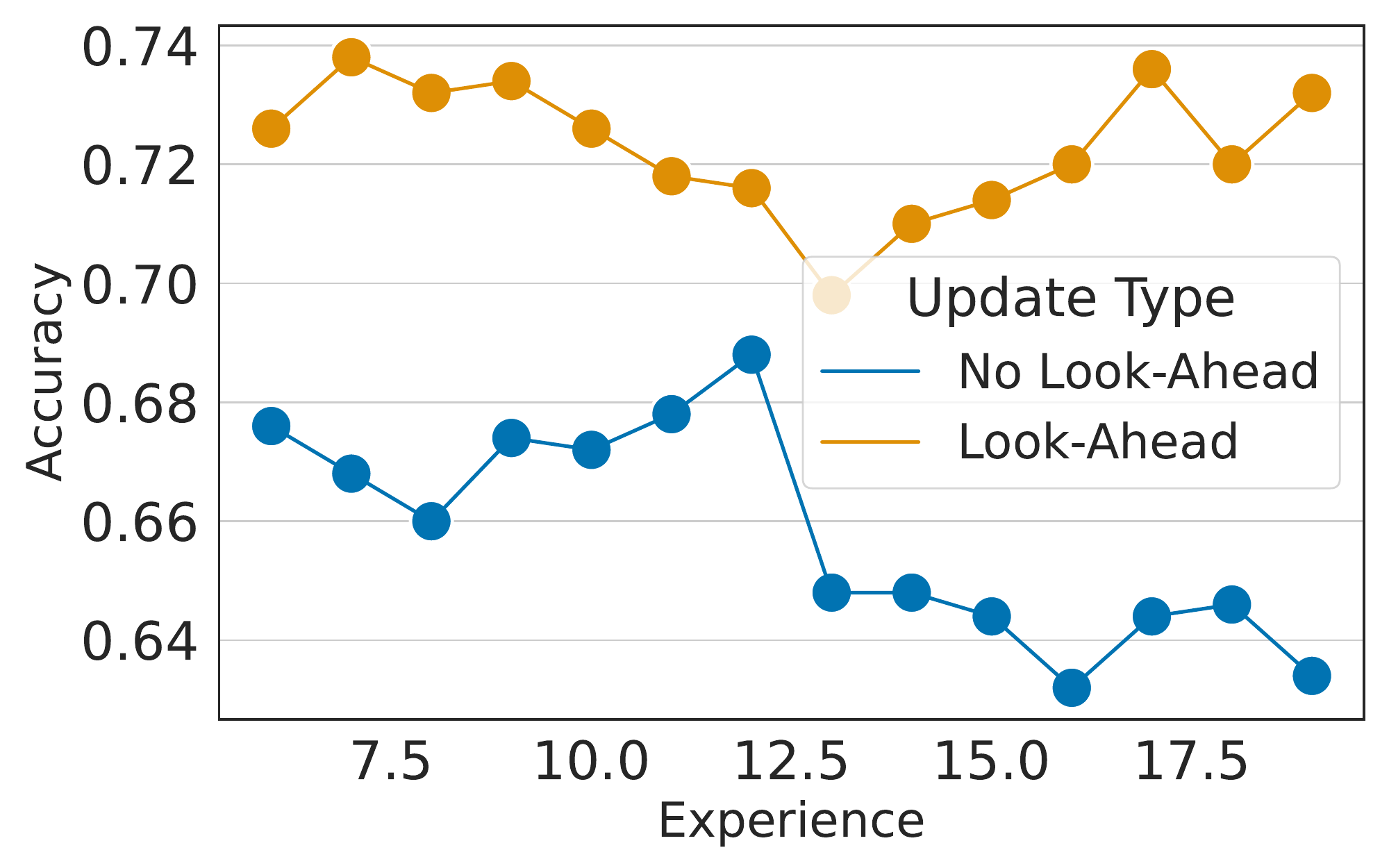}
    \includegraphics[width=0.24\linewidth]{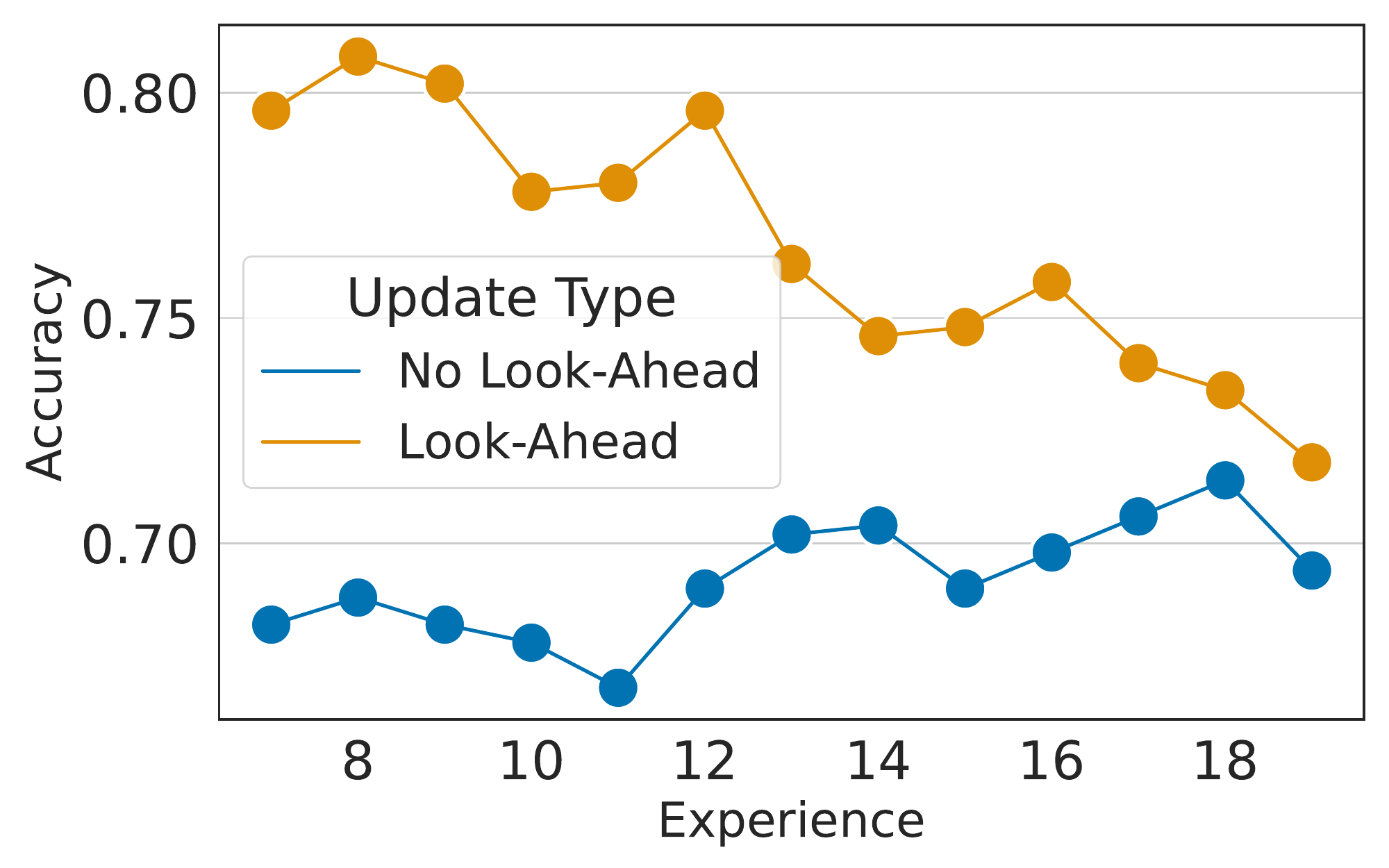}
   \caption{Test accuracy for experience $1$ (top-left) to experience $8$ (bottom-right) with and without look-ahead updates. The class order and model-initialization are the same for both modes.}
   \label{fig:lookahead}
   \vspace{-4mm}
\end{figure}

\begin{table}
  % \small
  \centering
  \begin{tabular}{l c c c c }
    \Xhline{2\arrayrulewidth}
    \textbf{Model} & \textbf{Update Type} & \textbf{ACA} & \textbf{Learning Accuracy} & \textbf{Running Time} \\
    \midrule
    \multirow{2}{*}{ResNet-$0$}  & No Look-Ahead & $62.7 $ & $65.0$ & $407$ Min  \\
                                 & Look-Ahead & $62.8 $ & $66.5$ & $636$ Min\\
    \midrule
    
    \multirow{2}{*}{ResNet-$1$}  & No Look-Ahead & $63.3 $ & $68.0$ & $377$ Min  \\
                                 & Look-Ahead & $61.5 $ & $66.2.0$ & $509$ Min\\
    \midrule

    \multirow{2}{*}{ResNet-$2$}  & No Look-Ahead & $61.0 $ & $64.2$ & $265$ Min \\
                                 & Look-Ahead & $63.2$ & $66.8$ & $389$ Min\\
    \midrule

    \multirow{2}{*}{ResNet-$3$}  & No Look-Ahead & $57.0 $ & $59.3$ & $165$ Min \\
                                 & Look-Ahead & $59.4 $ & $64.4$ & $243$ Min\\
    \midrule

    \multirow{2}{*}{ResNet-$4$}  & No Look-Ahead & $66.9 $ & $73.4$ & $87$ Min \\
                                 & Look-Ahead & $74.3 $ & $75.6$ & $115$ Min\\
    
    % \midrule
    \Xhline{2\arrayrulewidth}
  \end{tabular}
  \caption{Look-Ahead vs No Look-Ahead}
  \label{tab:look_ahead}
\end{table}

\subsection{First Order vs. Second-Order} \label{appendix_secondorder}
Since the computation of the second-order derivatives can be computationally very expensive, we compare both first-order and second-order based estimations of the gradients for the hypernetwork. As demonstrated in Table \ref{tab:secondorder}, for some model versions, the second-order estimation can be slightly better, while in others, the first-order can even perform better in terms of ACA. However, the running time for the second-order is significantly longer. Therefore, considering the huge gap in the running time, a first-order estimation is a more reasonable choice.

\begin{table}[h]
  % \small
  \centering
  \begin{tabular}{l c c c c }
    \Xhline{2\arrayrulewidth}
    \textbf{Model} & \textbf{Update Type} & \textbf{ACA} & \textbf{Learning Accuracy} & \textbf{Running Time} \\
    \midrule
    \multirow{2}{*}{ResNet-$0$}  & First-Order & $63.8$ & $66.5$ & $636$ Min \\
                                 & Second-Order & $64.1 $ & $66.2$ & $2406$ Min\\
    \midrule
    
    \multirow{2}{*}{ResNet-$1$}  & First-Order & $66.5$ & $66.2$ & $509$ Min  \\
                                 & Second-Order & $61.9 $ & $66.1$ & $1626$ Min\\
    \midrule

    \multirow{2}{*}{ResNet-$2$}  & First-Order & $63.2$ & $66.8$ & $389$ Min \\
                                 & Second-Order & $64.0 $ & $64.7$ & $1104$ Min\\
    \midrule

    \multirow{2}{*}{ResNet-$3$}  & First-Order & $59.4 $ & $64.6$ & $243$ Min \\
                                 & Second-Order & $56.6 $ & $61.9$ & $585$ Min\\
    \midrule

    \multirow{2}{*}{ResNet-$4$}  & First-Order & $74.3 $ & $75.6$ & $115$ Min \\
                                 & Second-Order & $73.1 $ & $75.0$ & $122$ Min\\
    
    % \midrule
    \Xhline{2\arrayrulewidth}
  \end{tabular}
  \caption{First Order vs. Second Order}
  \label{tab:secondorder}
\end{table}

\section{Memory Usage} \label{appendix_memorycomparison}

In this section we compare the memory usage between different versions of latent replay and partial hypernetworks in Table \ref{tab:memory_comparison}. For LR, the memory usage is calculated for the exemplars stored in the replay buffer, and for HN, the memory usage is equal to the size of the hypernetwork since we need to store the converged hypernetwok from the previous experience for computing the regularization term in the new experience. We assume $32$-bit tensors in our calculations.

The memory usage for exemplars in LR is affected by both increasing the size of the input and changing the architecture, in particular by changing the number of channels in a convolutional neural network. In HN, the memory usage is determined by the size of the main network, which can be reduced by partially generating the weights of the main model.
\renewcommand{\arraystretch}{1.1} 
\begin{table}[ht]
\centering
\caption{Memory usage comparison between different version of LR and HN. The buffer size for the CIFAR-100 and TinyImageNet datasets are $200$ and $400$ respectively.}
\label{tab:memory_comparison}
\renewcommand{\arraystretch}{1.2} % Adjust the spacing between rows

\begin{tabular}{c c c c c c c c}

\Xhline{2\arrayrulewidth}

& \multicolumn{3}{c }{\textbf{Latent Replay}} & & \multicolumn{3}{c }{\textbf{Partial Hypernetwork}} \\ 
& Model & Exemplar Shape & Memory Usage & & Model & \# of HN Parameters & Memory Usage \\

\cline{2-4} \cline{6-8} 
\multirow{5}{*}{\rotatebox[origin=c]{90}{CIFAR-100}} & LR-0 & $3\times 32 \times 32$ & $2.34$ MB & & HN-F & $1,272,877$ & $4.86$ MB \\
& LR-1 & $20\times 32 \times 32$ & $15.62$ MB & & HN-1 & $1,217,122$ & $4.64$ MB \\
& LR-2 & $4\times 16 \times 16$ & $7.81$ MB & & HN-2 & $1,119,522$ & $4.27$ MB \\
& LR-3 & $80\times 8 \times 8$ & $3.91$ MB & & HN-3 & $924,322$ & $3.53$ MB  \\
& LR-4 & $160\times 4 \times 4$ & $1.95$ MB & & HN-4 & $533,922$ & $2.04$ MB \\

\cline{2-4} \cline{6-8}  \\
\cline{2-4} \cline{6-8}  

% \multicolumn{3}{c }{\textbf{Latent Replay}} & & \multicolumn{3}{c }{\textbf{Partial Hypernetwork}} \\ 
& Model & Exemplar Shape & Memory Usage & & Model & \# of HN Parameters & Memory Usage \\
\cline{2-4} \cline{6-8}
\multirow{2}{*}{\rotatebox[origin=c]{90}{TinyImageNet}} & LR-0 & $3\times 64 \times 64$ & $18.75$ MB & & HN-F & $1,272,877$ & $4.86$ MB \\
& LR-1 & $20\times 64 \times 64$ & $125.0$ MB & & HN-1 & $1,217,122$ & $4.64$ MB \\
& LR-2 & $40\times 32 \times 32$ & $62.50$ MB & & HN-2 & $1,119,522$ & $4.27$ MB \\
& LR-3 & $80\times 16 \times 16$ & $31.25$ MB & & HN-3 & $924,322$ & $3.53$ MB \\
& LR-4 & $160\times 8 \times 8$ & $15.62$ MB & & HN-4 & $533,922$ & $2.04$ MB \\
&  & & &  & &  \\
\Xhline{2\arrayrulewidth}
\\

\end{tabular}
\end{table}

\section{Algorithm} \label{appendix_pseudocode}
In Algorithm \ref{alg:training_alg} we present the steps for training a partial hypernetwork over a CL stream.

\begin{algorithm}
\caption{Proposed algorithm for training a partial hypernetwork with look-ahead updates.}
\begin{algorithmic}[1]
\State \textbf{Input}: CL stream $\mathcal{S}$, learning rates $\alpha$ and $\beta$, regularization factor $\lambda$
\State Initialize parameters of the main model's stateful layers $\phi$
\State Initialize hypernetwork parameters $\omega$
\For{each experience $e_i$ in the CL stream $\mathcal{S}$}
    \If{$i = 1$}
        \Comment{Train with SGD updates}
        \While{not converged}
            \State Sample a mini-batch $b$ from $e_i$
            \State Compute gradients for $\phi$:  $g_{\phi} \gets \nabla_{\phi} \mathcal{L}(b, \{\phi, \omega\})$
            \State Compute gradients for $\omega$:  $g_{\omega} \gets \nabla_{\omega} \mathcal{L}(b, \{\phi, \omega\})$
            \State Update the main model's stateful parameters $\phi \gets \phi - \alpha g_{\phi}$
            \State Update hypernetwork parameters $\omega \gets \omega - \beta g_{\omega}$
        \EndWhile
        \State Freeze parameters $\phi$
    \Else
        \Comment{Update with look-ahead mechanism}
        \While{not converged}
            \State Sample a mini-batch $b$ from $e_i$
            \State Compute gradients $g_{\omega}^{(1)} \gets \nabla_{\omega} \mathcal{L}(b, \omega)$
            \State Make virtual update on $\omega$: $\omega^{\prime} \gets \omega - \beta g_{\omega}$
            \State Compute gradients using virtual parameters $\omega^{\prime}$: $g_{\omega}^{(2)} \gets \nabla_{\omega} \mathcal{L}(\mathcal{B}, \omega^{\prime})$
            \State Compute the average of $g_{\omega}^{(1)}$ and $g_{\omega}^{(2)}$: $\overline{g}_{\omega} =\frac{g_{\omega}^{(1)}+g_{\omega}^{(2)}}{2}$ 
            \State Update hypernetwork parameters $\omega \gets \omega - \beta \overline{g}_{\omega}$
        \EndWhile
    \EndIf
\EndFor
\end{algorithmic}
\label{alg:training_alg}
\end{algorithm}

\section{Gradient Similarity} \label{appendix_hyerparameters}
\begin{figure}[h!]
    % \vspace{-4mm}
  \centering
%   \fbox{\rule{0pt}{2in} \rule{0.9\linewidth}{0pt}}
    \includegraphics[width=0.4\linewidth]{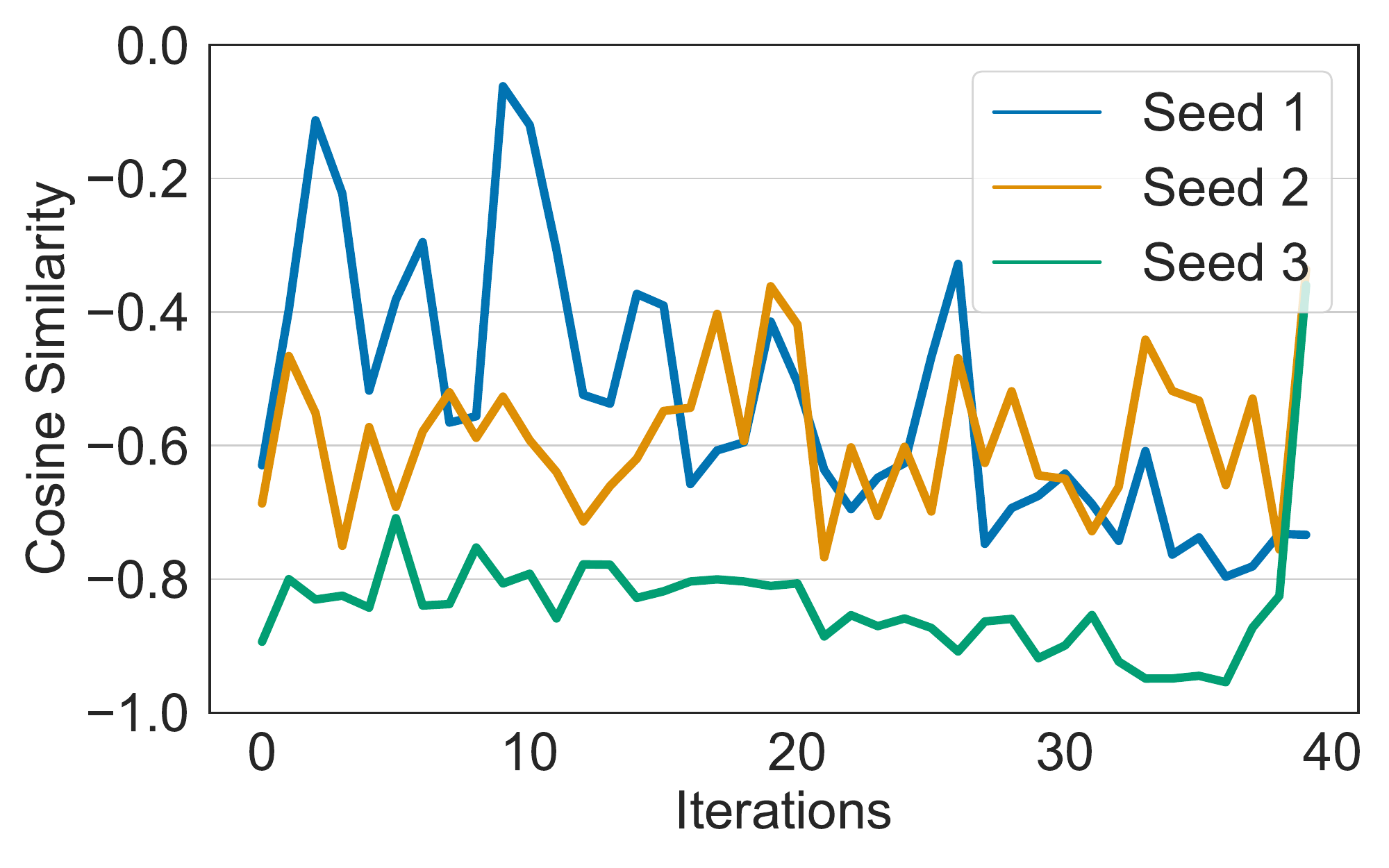}
   \caption{Cosine similarity between gradients $g_{\omega}^{(1)}$ and $g_{\omega}^{(2)}$.}
   \label{fig:gradient_conflict}
   % \vspace{-4mm}
\end{figure}

In Figure \ref{fig:gradient_conflict} we demonstrate the cosine similarity of the hypernetwork gradients $g_{\omega}^{(1)}$ (from the cross-entropy loss) and $g_{\omega}^{(2)}$ (from the virtual update) in HN-4, as computed in Algorithm \ref{alg:training_alg} in the first epoch of the second experience. The plots show negative similarity regardless of the class order or the model initialization which are affected by the random seeds.

\section{Hyperparameter Selection} \label{appendix_hyerparameters}
We carried out a hyper-parameter selection at the stream level. For common hyper-parameters like learning rate, we implemented a grid search and assessed the model's final performance on the test set of each stream across three runs (using 3 random seeds). In our experiments, we used a learning rate of $0.001$ for both models.

For the hypernetwork, we executed a grid-search, opting for the minimum embedding size capable of attaining a similar performance. This approach allowed us to minimize the hypernetwork's number of parameters. In our experiments, we used dimension sizes $32$, $50$, $32$ and $32$ for the embedding dimension the three consecutive layers of the multi-layer peceptron respectively. We set the hypernetwork's regularization coefficient to $0.5$.

\end{document}